\documentclass[10pt,twocolumn,letterpaper]{article}

\usepackage{times}
\usepackage{epsfig}
\usepackage{graphicx}
\usepackage{amsmath}
\usepackage{amssymb}
\usepackage{booktabs}
\usepackage{comment}
\usepackage{amsthm}
\usepackage[toc,page]{appendix}
\usepackage{pifont} 
\usepackage{arydshln}
\usepackage{multirow}
\usepackage[pagebackref,breaklinks, colorlinks]{hyperref}
\usepackage{bookmark}
\usepackage[pagenumbers]{wacv}
\def\wacvPaperID{1001} 
\def\confName{WACV}
\def\confYear{2024}

\usepackage[capitalize]{cleveref}
\crefname{section}{Sec.}{Secs.}
\Crefname{section}{Section}{Sections}
\Crefname{table}{Table}{Tables}
\crefname{table}{Tab.}{Tabs.}

\begin{document}

\title{TIAM - A Metric for Evaluating Alignment in Text-to-Image Generation}

\author{\begin{tabular}[t]{c@{\extracolsep{1.2em}}c@{\extracolsep{1.2em}}c@{\extracolsep{1.2em}}c}
    Paul Grimal &
    Hervé Le Borgne
                & Olivier Ferret &
    Julien Tourille                \\
\end{tabular}
{} \\
\\
Université Paris-Saclay, CEA, List, F-91120, Palaiseau, France
    {} \\
{\tt\small \{paul.grimal, herve.le-borgne, olivier.ferret, julien.tourille\}@cea.fr}
}

\maketitle

\begin{abstract}
    The progress in the generation of synthetic images has made it crucial to assess their quality. While several metrics have been proposed to assess the rendering of images, it is crucial for Text-to-Image (T2I) models, which generate images based on a prompt, to consider additional aspects such as to which extent the generated image matches the important content of the prompt. Moreover, although the generated images usually result from a random starting point, the influence of this one is generally not considered. In this article, we propose a new metric based on prompt templates to study the alignment between the content specified in the prompt and the corresponding generated images. It allows us to better characterize the alignment in terms of the type of the specified objects, their number, and their color. We conducted a study on several recent T2I models about various aspects. An additional interesting result we obtained with our approach is that image quality can vary drastically depending on the noise used as a seed for the images. We also quantify the influence of the number of concepts in the prompt, their order as well as their (color) attributes. Finally, our method allows us to identify some seeds that produce better images than others, opening novel directions of research on this understudied topic.
\end{abstract}

\section{Introduction}

The ability to generate synthetic images with neural models made significant advancements from the advent of the first GANs~\cite{goodfellow2014gan,radford2016dcgan}. More recently diffusion-based models ~\cite{ho2020denoising,rombach2021highresolution,saharia2022photorealistic,nichol2022glide,balaji2023ediffi} have further pushed the boundaries of image synthesis by progressively denoising an initial noise to generate high-quality images. In parallel to these advances, the question of evaluating the quality of these synthetic images has always been a delicate issue and has become a research question in itself. To address this problem~\cite{theis2016eval_generative}, several metrics were proposed~\cite{heusel2017fid,salimans2016improved_gans,zhang2018lpips} but they suffer from various limits~\cite{barratt2018note_is,borji2019pro_cons_metrics_generative}.
The most recent models are conditioned on textual image descriptions, allowing fine control of the output. It nevertheless adds a challenge to evaluate their outputs, namely to estimate to which extent the synthetic image generated corresponds to the textual description it was conditioned on.

Although Text-to-Image (T2I) models demonstrate strong semantic and compositional capabilities, achieving a visually pleasing image that aligns with the desired condition often requires the generation of multiple images to obtain a suitable one. A reliable generative model should exhibit alignment with the condition specified in the prompt, irrespective of the starting noise. To address and study the variability of results, we introduce a novel metric to assess the success rate of generative models according to a prompt, Text-Image Alignment Metric\footnote{Source code: \href{https://github.com/grimalPaul/TIAM}{https://github.com/grimalPaul/TIAM}} (TIAM). The initial noise plays a crucial role in our metric, enabling us to investigate its impact. We show that certain initial noise configurations outperform others, suggesting the possibility of selecting them to get better synthetic images.

\begin{figure}[ht]
    \centering
    \begin{tabular}{ccc}
        \includegraphics[width=0.28\linewidth]{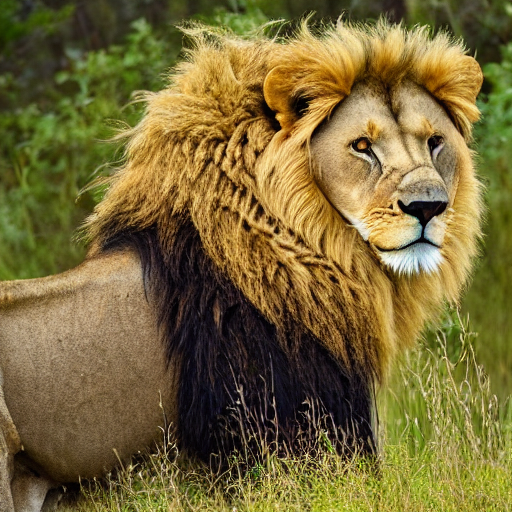}             &
        \includegraphics[width=0.28\linewidth]{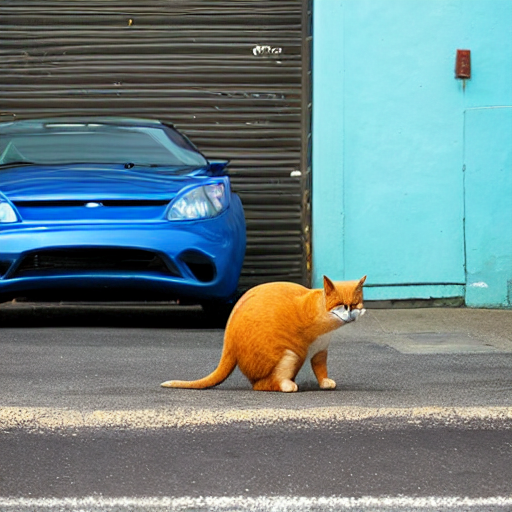} &
        \includegraphics[width=0.28\linewidth]{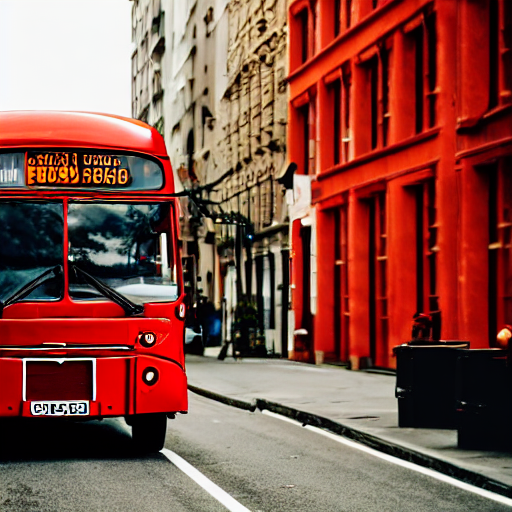}                           \\
        (a) Catastrophic                                                                                     & (b) Attribute & (c) Attribute \\
        neglect                                                                                              & binding       & leaking       \\
    \end{tabular}
    \caption{Images generated with the prompts ``\textit{a photo of a lion and a bear}'', ``\textit{a photo of a blue cat and a yellow car}'', and ``\textit{a photo of a red bus driving down the street}'' generated with Stable diffusion v1.4. (a) The bear is missing, (b) the attributes are swapped, (c) the bus color (\textit{red}) leaks on the wall}
    \label{fig:ex_issues}
\end{figure}

Recent research efforts \cite{chefer2023attendandexcite,ramesh2022hierarchical,saharia2022photorealistic,tang2022daam, feng2023trainingfree} have shown that text-conditioned diffusion models suffer from three main issues related to the alignment between the expected content expressed in the textual prompt and the one actually generated in the image: (i) \textit{catastrophic neglect}, where one or more elements described in the prompt are not generated or sometimes mixed, (ii) \textit{attribute binding}, where attributes (\eg color) are bound to the wrong entities, and (iii) \textit{attribute leaking}, where attributes specified in the prompt are correctly bound but some other elements in the scene are also wrongly bound with this attribute (Fig.~\ref{fig:ex_issues}).

With TIAM, we propose to analyze the success rate of generative models under the scope of catastrophic and attribute-binding issues. For the latter, we propose a solid method to evaluate color alignment with human perception.

To date, the investigation of the influence of some words and attributes remains largely understudied. Tang \etal \cite{tang2022daam} provide some insights by examining the impact of words on generated outcomes. Using TIAM, we provide further insights into the relationship between textual conditioning and generative results. Relying on prompt templates rather than natural prompts, TIAM allows quantifying the prompt-image alignment w.r.t syntactic aspects, in particular the importance of the position of the main entities in the prompt.

In summary, our main contributions are:
(i) a new metric based on prompt templates to quantify automatically the performance of T2I models \textit{in terms of prompt-image alignment}; (ii) an in-depth study of several diffusion models and \textit{quantification} of their behavior relating to catastrophic neglect and attribute binding issues; (iii) a study on the influence of the initial diffusion noise (seed) for all these models. The main insights resulting from this study are (a) the alignment performance of most T2I models drops significantly with the number of objects specified in the prompt; (b) in practice there exists some seeds that systematically lead to better results; TIAM allows us to identify them and they still provide better results with objects that are out of our study domain; (c) most T2I models succeed in attributing color to one object but the performance drop with more objects.

\section{Related Work}

\paragraph{Text-to-image models}
Among the current state-of-the-art models, diffusion models \cite{ho2020denoising} have demonstrated remarkable performance. These models introduce noise to images and learn to denoise the added degradation. During inference, the models iteratively denoise a noise sampling from a Gaussian distribution, resulting in a reconstructed image. The utilization of free classifier guidance \cite{ho2022classifierfree} allows users to express their thoughts by writing a prompt using natural language. The diffusion process is then guided by the encoded text representation, leveraging foundational models like CLIP \cite{radford2021learning} or T5 \cite{raffel2020exploring}. It is important to note that the results are highly stochastic, often requiring the generation of multiple images with different Gaussian noise inputs to align with the user's desired texts and preferences. Among the well-known models in this field, notable mentions include Imagen~\cite{saharia2022photorealistic}, Dall-E~2~\cite{ramesh2022hierarchical}, Latent Diffusion Model (LDM)~\cite{rombach2021highresolution} and Ediff-I~\cite{balaji2023ediffi}.

\paragraph{T2I Evaluation}

The evaluation of generated images often relies on metrics such as Inception Score (IS) \cite{salimans2016improved_gans} and Fréchet Inception Distance (FID) \cite{heusel2017fid}, which are commonly used to assess their quality and fidelity. However, these metrics do not capture the alignment between the provided condition(s) and the generated images. To address this issue, several methods have been proposed to measure whether the content of the generated images reflects the given conditions. One approach involves utilizing classification models \cite{ravuri2019classification} and object detection models (Semantic Object Accuracy, abbreviated as SOA) \cite{HinzSOA2022} to determine whether the generated images contain the specified objects or adhere to certain criteria. SOA uses actual prompts from COCO while our approach relies on prompt templates that allow a finer analysis of the influence of each element in the prompt. Another method specific to text-image alignment involves employing visual score similarity measures. It leverages contrastive models such as CLIP, which compute a similarity score between representations of images and text. Although CLIP has a great average semantic representation, it has poor compositional understanding \cite{yuksekgonul2023when}, limiting a fine evaluation of text-image alignment. Our metric addresses this issue by leveraging a high-performance detector and by controlling the requested attributes in an explainable manner and aligning with human perception.

Recent efforts have introduced innovative methodologies to assess skills and measure biases of T2I models. Notably, Drawbench \cite{saharia2022photorealistic} presents a limited set of text prompts covering 11 skills to evaluate the models. Similarly, DALL-EVAL \cite{cho2022dalleval} proposes to evaluate three visual reasoning skills: the capacity to generate one object (object recognition), the capacity to generate the exact quantity number of the asked object (counting), and the capacity to place two objects (spatial reasoning). Additionally, they probe gender and skin bias in the model representation. Another contribution by Zhang \etal \cite{zhang2023auditing} delves into the gender depiction disparities enabling the study of potential stereotypes. Our methods focus on catastrophic neglect and attribute binding. We can see the catastrophic neglect approach as object recognition but by testing one or more different objects in the conditioning prompt. Moreover, to overcome results that are very sensitive to asked objects in prompt and evaluate the model's performance accurately systematically, we explore all various object and attribute combinations and we generate multiple images per prompt to enhance result robustness. Recognizing the impact of initial noises, we advocate for multiple seed testing. This approach offers a comprehensive evaluation, free from the limitations of specific labels, resulting in a more precise and less biased assessment. A closely related work \cite{gokhale2022benchmarking} conducted concurrently with ours shares some conclusions, though it does not delve into the aspect of the starting noises.

\section{Method}

We propose a new method to measure the success rate of generative models w.r.t catastrophic neglect and/or attribute binding. First, we generate multiple prompts given a set of word labels and possible attributes. We then generate multiple images for each prompt and detect if the expected elements are present on the image, leading to the final score.

\subsection{TIAM Text-Image Alignment Metric \label{sec:3_1_TIAM}}
Our approach is based on templates that are used to generate sets of prompts. We adopt a formalism inspired by the disentangled representation theory~\cite{higgins2018disentangled_repr,trager2023linear}. The prompts contain $N$ objects, each of which can be qualified by an attribute. Hence the object at position $i$ in a prompt is a token belonging to the set $\mathcal{O}_i$ and qualified by an attribute in the set $\mathcal{A}_i$. For example, let us consider the template ``\textit{a photo of \texttt{det}($o_1$, $a_1$) $a_1$ $o_1$ and \texttt{det}($o_2$, $a_2$) $a_2$ $o_2$}'' where $o_i\in\mathcal{O}_i$, $a_i\in\mathcal{A}_i$, and \texttt{det}($o_i$, $a_i$) is a determinant that depends on the object or attribute if present. If $\mathcal{O}_1=\mathcal{O}_2=$\{`car',`bike',`truck'\} and $\mathcal{A}_1=\mathcal{A}_2=$\{`blue',`green',`red'\}, then it can produce prompts such as ``\textit{a photo of a blue car and a red truck}'' or  ``\textit{a photo of a green bike and a blue car}''.

For such a template $t$, the generic expression of the text-image alignment metric (TIAM) is defined as:
\begin{equation}\label{eq:tiam}
    \underset{\substack{\chi\sim\mathcal{N}(0,I)\\ z\in \mathcal{Z}}}{\mathbb{E}} \left[ f\left(G(\chi, t(z)),y(z)\right)\right] \quad \text{with:} \quad
    \mathcal{Z}  =  \prod_{i=1}^N\left(\mathcal{A}_i\times\mathcal{O}_i \right)
\end{equation}
where the prompt is instanced from the template $t$ and its ``latent concepts'' $z\in\mathcal{Z}$, $y(z)$ is the labels that relate to the expected content of the synthetic image generated by the model $G()$ (conditioned by the prompt) from a \textit{seed} (using to generate an initial noise) $\chi$.  $f$ is a scoring function that compares the ground truth $y(z)$ to the output of a model that detects objects or produces segmentation maps from the synthetic image, the resulting score being in $\{0,1\}$, depending on whether the content matches the ground truth or not (see Section~\ref{sec:impl_detail}).

With the definition of \autoref{eq:tiam}, a template can generate $\sum_{i=1}^N|\mathcal{A}_i|.|\mathcal{O}_i|$ different prompts, where $|.|$ is the cardinal of the set. However, we restrict in practice the inference of the template such that an object or an attribute can not appear twice in the prompt. In a simple case where the attribute and  object sets are the same at each position, the number of unique prompts is given by:
\paragraph{Proposition 1:} if $|\mathcal{O}|\ge N$, $|\mathcal{A}|\ge N$, and $\forall i\in [\![1,N]\!], \mathcal{A}_i=\mathcal{A}, \mathcal{O}_i=\mathcal{O}$, and $\forall (i,j)\in [\![1,N]\!]^2$ s.t $i<j$, we force $a_i\ne a_j$ and $o_i\ne o_j$, thus the number of unique prompts generated in the context of \autoref{eq:tiam} is $\frac{|\mathcal{O}|!|\mathcal{A}|!}{\left(|\mathcal{O}|-N\right)!\left(|\mathcal{A}|-N\right)!}$

\begin{figure}
    \centering
    \includegraphics[width=1\linewidth]{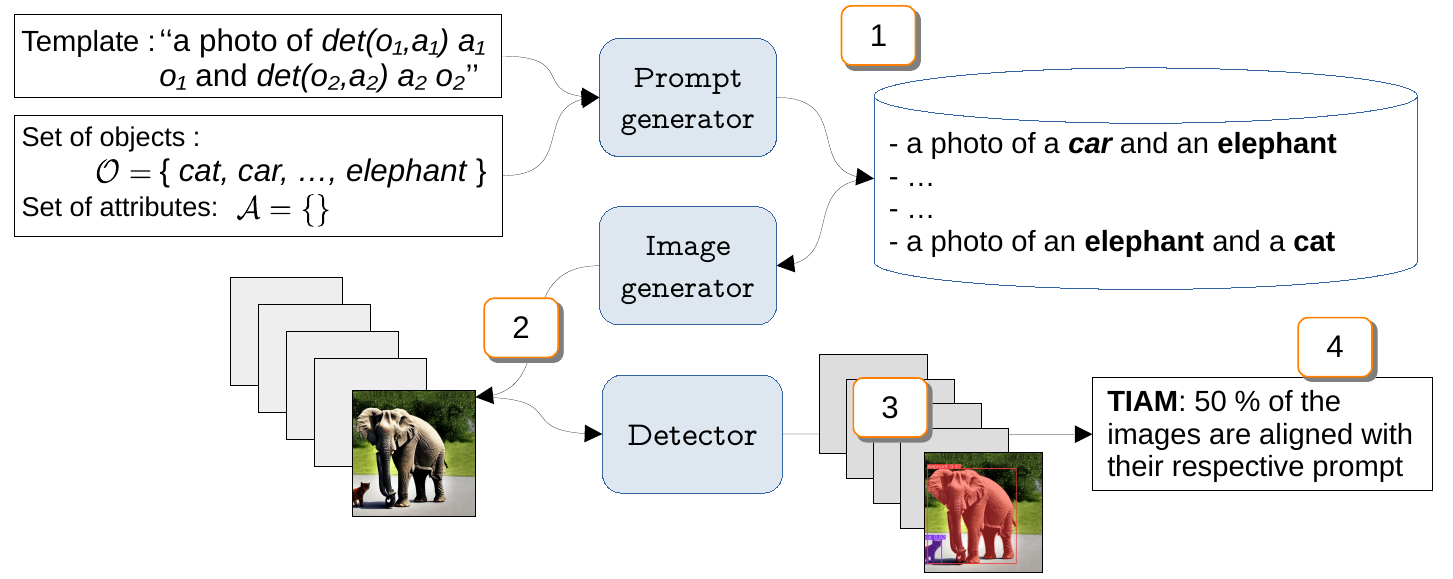}
    \caption{Overview of the evaluation pipeline. (1) Generate a dataset of prompts. (2) Generate $n \ge 16$ images per prompt. (3)~Detect if the requested labels are present in the images. (4) Compute TIAM. In this example, we do not define attributes.}
    \label{fig:pipeline_eval}
\end{figure}

The proof is given in the Supp. Mat.~\ref{sec:proof_permutation} (supplementary material). We also derive a more general case with possibly different attributes and objects at each position. The global evaluation process is summarized in Fig.~\ref{fig:pipeline_eval}.

\subsection{Attribute definition\label{sec:attribute_definition}}

We study the color attribute in this paper, but our metric could be applied under the scope of another type of attributes (\eg, size or texture). Selecting colors that are aligned with human perception is not trivial because of the infinity of possibilities. We base our choice on the work of Berlin and Kay \cite{BerlinBasicColor1969}. They define eleven universal \textit{basic colors} $\mathcal{C}$: \textit{white, black, red, green, blue, brown, purple, pink, orange, yellow}, and \textit{grey}. They asked individuals to select from an array of 329 colors (provided by the Munsell Colors Company) the chips that correspond to each \textit{basic color} and to select the most typical examples. We use the results of the most typical examples of American English and convert the colors from the Munsell System to the CIELAB space. This one offers the advantage of a color distribution more aligned with human perception and a suitable space to compare color differences.
We remarked that \textit{brown} and \textit{orange} were respectively too close to \textit{black} and \textit{white} and we removed them (more information provided in Supp. Mat.~\ref{appendix:other_attributes}). We set $\mathcal{A}=\{\mathit{red, green, blue, purple, pink, yellow}\}$. However, to determine the actual colors of the pixels in the CIELAB space, we also consider \textit{white} and \textit{black}.

To define if $G()$ correctly assigns $a_i$ and $o_i$, we use a segmentation model that delimits $o_i$ if present on the image. We compare all the pixels inside the segmentation maps with our reference colors in the CIELAB space. If we detect at least a proportion of 40\% of $a_i$, we consider a successful binding.

\subsection{Main implementation details}\label{sec:impl_detail}

The template $t$ always includes the start mention ``\textit{a photo of}''. Indeed, in our case, the detector is trained on real images and we suppose that it would therefore have more difficulty managing \eg sketches or 3D renders. The general form of the templates is that given as an example above, by varying the syntactic context.

For the detection and segmentation tasks, we use  YOLOv8~\cite{yolov8_ultralytics}, a state-of-the-art model pre-trained on the 80 COCO labels \cite{lin2015microsoft}. We set the confidence threshold to 0.25 but conducted a study asserting that a higher threshold (0.4 to 0.8) does not change the relative order performance of the models. The results can be found in the Supp. Mat.~\ref{appendix:detailSegmentationDetection}, with a detailed description of the labels used in each experiment and their combination within the generated prompts.

The scoring function $f$ mainly focuses on true positives, considering the alignment as successful if we find at least one of each named object in the prompt. If the objects are characterized by an attribute, each object must be present with its correct attribute at least once in the image to be marked as a success. The criterion is therefore strict on the alignment aspect, requiring the presence of \textit{all} the named objects and attributes. This approach differs from CLIP scores, which provide a general trend without reflecting a comprehensible explanation of the results. It nevertheless does not penalize the presence of other objects (false positive) since it is not prohibitive in a generative context. Indeed, if the segmentation masks of two different objects overlap with an IoU value of 0.95 or higher, we remove both objects before calculating the score with the remaining detected objects.

\section{Study}
We conduct an in-depth study to characterize and quantify the limits of several diffusion models, which differ by the text-conditioning method, the architecture, or the inference process. We consider Stable Diffusion v1-4 (SD 1.4) and Stable Diffusion v2 (SD 2) based on the LDM model~\cite{rombach2021highresolution} that uses a fixed pre-trained text encoder CLIP~\cite{radford2021learning}, which is based on an auto-regressive architecture. Both versions are trained with the same variational autoencoder (VAE), but two different U-nets (the part that learns to denoise image). The CLIP models that guide the U-nets also differ between SD~1.4 and SD~2. In addition, we evaluate the two SD models with Attend-and-Excite (A\&E)~\cite{chefer2023attendandexcite}, an optimized inference process that uses the cross-attention maps to attend the subject tokens in the prompt. We follow the authors' implementation and only attend to the token of object~$o_i$, even if an attribute characterizes the object. We also consider an unCLIP model~\cite{ramesh2022hierarchical,kakaobrain2022karlo-v1-alpha} conditioned with a CLIP image prior and a CLIP text embedding. Finally, we study DeepFloyd IF (IF), a cascade diffusion inspired by~\cite{saharia2022photorealistic}, conditioned using a T5 XXL \cite{raffel2020exploring} text encoder. Other details are reported in the Supp. Mat.~\ref{appendix:setupModelT2I}.

\subsection{Preliminary: performance drop with 2 objects}\label{sec:preliminary}
We report a first experiment to illustrate the general framework of our study. We consider various labels from COCO (the exhaustive list is in Supp. Mat.~\ref{appendix:prompt_template}), leading to a set of $|\mathcal{O}|=24$ object labels. For all possible pairs of labels $(o_i,o_j)$ with $o_i\ne o_j$, we make the prompt ``a photo of $o_i$ and $o_j$'' (managing the determinant as expected) and generate 64 synthetic images by changing the random seed $\chi$ and estimate the alignment of the prompt with each image using TIAM. As a reference, we also conduct the same experiment with 64 images generated from the simpler prompt ``a photo of $o_i$''.

We report the results in Tab.~\ref{tab:benchmark_24} for all the models considered. The models consistently succeed in generating images with a single object, with a score above 0.95. However, they struggle to generate simultaneously two objects, with at most 66\% of images correctly generated, while the prompts are minimally simple. In the recent literature, the T2I models usually exhibit a good quantitative rendering score, \eg in terms of inception score or FID, which our experience does not call into question. In the vein of Hinz \etal~\cite{HinzSOA2022}, this experiment only shows that there is a problem with alignment between the content of the generated image and the prompt. More precisely than~\cite{HinzSOA2022}, our experiment specifically quantifies to which extent this drop in alignment performance is due to the presence of multiple concepts in the prompt. For the prompts with two objects, we report in Tab.~\ref{tab:appearance_object} the number of times the objects are actually accurately generated with regard to their position in the prompt (the score in Tab.~\ref{tab:benchmark_24} requires the presence of both objects to be valid). We observe a tendency for the first $o_1$ to be more prevalent than the second $o_2$ across all models, a phenomenon that is further investigated in Section~\ref{sec:catastrophic_neglect}.

We notice that A\&E inference globally improves the score with a large margin for SD v1.4. From Tab.~\ref{tab:appearance_object}, it seems that the overperformance of A\&E is due to its ability to enhance the occurrence rate of $o_2$ by enforcing the minimum excitation of cross-attention maps.

\begin{table}[]
    \centering
    {\small{
            \begin{tabular}{lrr}
                \toprule
                Model       & 1 object & 2 objects \\
                \midrule
                SD 1.4      & 0.98     & 0.41      \\
                SD 1.4 A\&E & 0.96     & 0.64      \\
                SD 2        & 0.99     & 0.61      \\
                SD 2 A\&E   & 0.98     & 0.65      \\
                unCLIP      & 0.95     & 0.50      \\
                IF          & 0.99     & 0.62      \\
                \bottomrule
            \end{tabular}
        }}
    \caption{TIAM according to the number of objects in the prompt, for all 6 generative models considered in the study.}
    \label{tab:benchmark_24}
\end{table}

\begin{table}[]

    \centering
    {\small{\begin{tabular}{lll}
                \toprule
                Model       & $o_1$ & $o_2$ \\
                \midrule
                SD 1.4      & 0.80  & 0.60  \\
                SD 1.4 A\&E & 0.85  & 0.75  \\
                SD 2        & 0.83  & 0.78  \\
                SD 2 A\&E   & 0.84  & 0.80  \\
                unCLIP      & 0.77  & 0.71  \\
                IF          & 0.86  & 0.76  \\
                \bottomrule
            \end{tabular}}}
    \caption{Proportion of appearance per order in the prompt. $o_1$ and $o_2$ refer to the position in the template.}
    \label{tab:appearance_object}
\end{table}

\subsection{Importance of the random seed}\label{sec:random_seed}
We consider the same 24 labels as in Section~\ref{sec:preliminary} and the prompt ``a photo of $o_i$ and $o_j$''. We also generate 64 synthetic images by changing the random seed $\chi$ and compute $f\left(G(\chi, t(z)),y(z)\right)$ for each prompt and the corresponding image. However, this time, the 64 seeds are the same for all the $(o_i,o_j)$\footnote{Actually, it was the case in Section~\ref{sec:preliminary} for the sake of comparison and coherency.} and we aggregate the performance \textit{per seed}, for all possible prompts and images, leading to  $24\times 23=552$ estimations of text-image alignment per seed.

\begin{figure}
    \centering
    \includegraphics[width=0.7\linewidth]{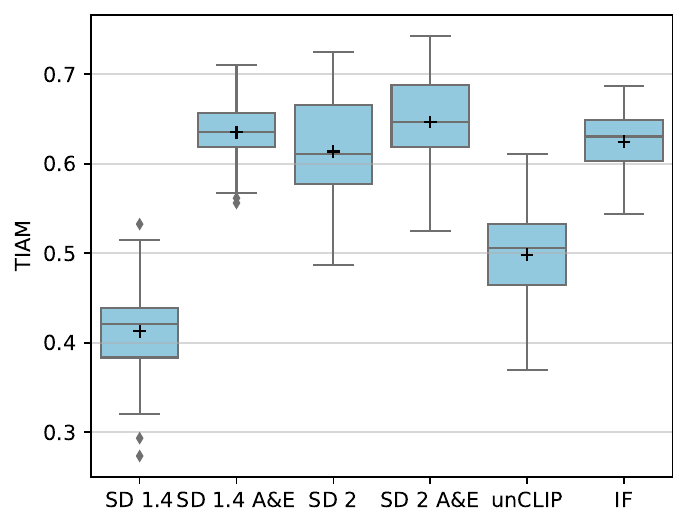}
    \caption{TIAM aggregated per seed for 64 seeds. We show that some starting noise tends to not be converted to an image with two entities regardless of the entities. ``+'' shows the mean.}
    \label{fig:boxplot_TIAM_seed}
\end{figure}

The results in Fig.~\ref{fig:boxplot_TIAM_seed} show that, for all 6 models, the performance varies a lot with regard to the random seed. These results, which to the best of our knowledge have never been identified in the literature of the T2I generative models, may seem surprising. With such models, one usually expects that all seeds have the same chance to generate the specified elements. Actually, the recent tendency rather consists of engineering finely the prompt~\cite{oppenlaender2023taxonomy} to optimize the output of T2I models. On the opposite of this tendency, our method opens up new perspectives toward a complementary optimization based on the choice of ``performing seeds'' (see Section~\ref{sec:illustrative_prompts}).

Overall, there exists a significant disparity between the best and worst seeds. When considering the interquartile ranges and the min-max range through the box-plots of Fig.~\ref{fig:boxplot_TIAM_seed}, it appears that the difference in average performance between models is much less significant than their own inner difference due to the variation of the seeds.

Obviously, enhancing a dependence ``to the seed'' is convenient from a practical point of view but wrong in all strictness since the initial noises are drawn from a Gaussian at inference. Being more rigorous requires reminding the diffusion models training process~\cite{ho2020denoising}. It consists of learning the reverse process of a fixed Markov chain of length $T$ with models that can be interpreted as an equally weighted sequence of denoising autoencoders. The latters are trained to denoise their input, considered as a noisy version of an input training image $I$. In the case of latent diffusion models~\cite{rombach2021highresolution}, the process is embedded into the latent space of a VAE (encoder $\mathcal{E}$ + decoder $\mathcal{D}$), such that the autoencoders are U-net networks $\epsilon_\theta(x_t,t)$, $t=1\dots T$ that denoise the latent code $x_t$ by minimizing the loss $\mathbb{E}_{\mathcal{E}(I),\epsilon\sim\mathcal{N}(0,1)}\left[ || \epsilon -\epsilon_\theta(x_t,t)||_2^2\right]$. During inference, the process starts from a latent code $\chi_T\sim\mathcal{N}(0,1)$, denoises it with the U-net $\epsilon_\theta$ to get the final latent code $\chi_0$, then obtains the synthetic image with VAE decoder as $\mathcal{D}(\chi_0)$. This last is thus not just dependent on the random seed $\chi_T$ but on the U-net parameter $\theta$ as well. These parameters are learned to denoise the latent codes at any step of the diffusion since the reparametrization trick~\cite{ho2020denoising} allows to express the final code directly as:
\begin{equation}\label{eq:reparam_trick}
    x_0 = \frac{x_t - \sqrt{1-\bar{\alpha_t}}\epsilon}{\sqrt{\bar{\alpha_t}}}
\end{equation}
where $\epsilon\sim\mathcal{N}(0,1)$ and the $\alpha_t$ depends on the variance of the diffusion process noise (see~\cite{ho2020denoising} and  appendix B of \cite{rombach2021highresolution} for details). 
Ideally, $\bar{\alpha_T}$ is close to one, such that a synthetic image should result from an almost perfect Gaussian noise. However, in practice the process is not perfect; thus the final $\chi_0$ obtained for a given $\chi_T\sim\mathcal{N}(0,1)$ at inference time is only an approximation of the ``ideal $\chi_0$'' that could be expected with a perfect optimization~\cite{dietterich2000ensemble_methods}.
Moreover, since each model is trained independently, its latent space is structured in the same vein and the ``path'' of the $\chi_t$ in the latent space may strongly differ from one model to another. Hence, starting from the same $\chi_T\sim\mathcal{N}(0,1)$, they result in quite different $\chi_0$ while trying to tend to an ideal one.  
As a consequence, the ``good'' and ``bad'' seeds are specific to each model.

The results depend on the seed, indicating that based on the prompt's specifications, it appears possible to identify initial noises that are more likely to exhibit multiple objects. It highlights the ongoing need for further advancements in reducing reliance on latent variables. To obtain a robust score and avoid the possible impact of the high variability of the seed's success rate, we define a minimum of images to generate per prompt. We establish it is adequate to generate 32 images per prompt to obtain a robust score. The corresponding experiment is reported in the Supp. Mat.~\ref{appendix:minImgGenerate}.

\subsection{Catastrophic Neglect}\label{sec:catastrophic_neglect}
\begin{figure}[ht]
    \centering
    \includegraphics[width=0.8\linewidth]{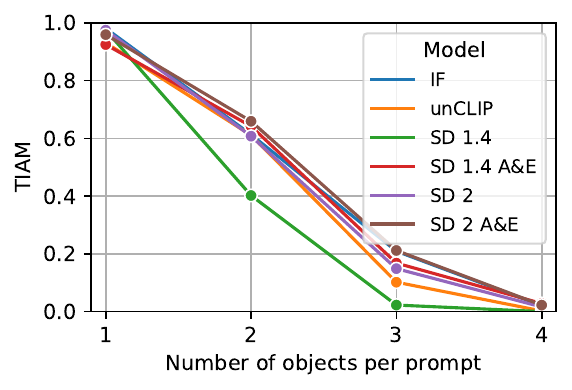}
    \caption{TIAM with 1 to 4 objects per prompt.}
    \label{fig:TIAM_nb_entity}
\end{figure}

We study the behavior of the models as the number of object sets increases and the role of the order of objects in the prompt. We set $\mathcal{O}=\{$\textit{car, refrigerator, giraffe, elephant, zebra}$\}$ and compute TIAM for prompts containing from 1 to 4 objects. We design 4 templates, one per number of objects in the prompt, and generate 32~images for each prompt, leading to 160~alignment values, 640~values, 1,920~values, and 3,820~values respectively with one, two, three, and four objects in the prompt (for each model).
As shown in Fig.~\ref{fig:TIAM_nb_entity}, the models fail to consistently generate outputs when prompted with more than two objects. Even with the A\&E mechanism, generating four objects remains nearly impossible. SD 2 and IF demonstrate relatively better performance, but the improvement is not substantial.

\begin{figure}[ht]
    \centering
    \includegraphics[width=0.8\linewidth]{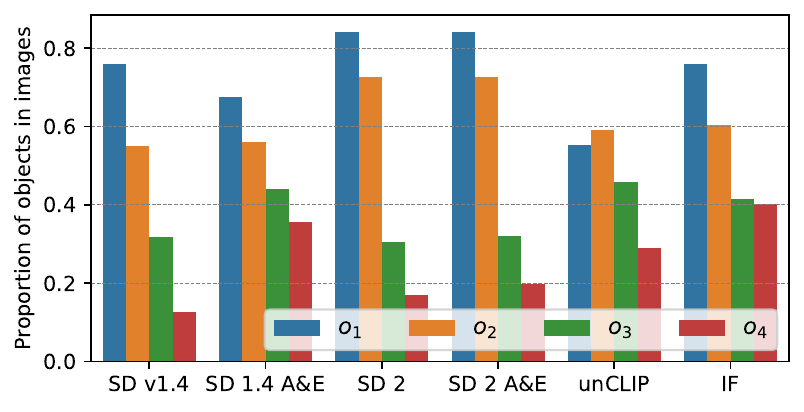}
    \caption{The proportion of occurrences of each object, based on its position in the prompt (template with 4 objects).}
    \label{fig:apparition_4}
\end{figure}

When examining the occurrence of different objects, similar trends can be observed as in the preliminary experiment (Tab.~\ref{tab:appearance_object}). Specifically, the initial objects in the template tend to appear more frequently than objects inserted subsequently. The results for the template containing four objects are presented in Fig.~\ref{fig:apparition_4} (results for two and three objects in Supp. Mat.~\ref{appendix:occurenceOfObject}). This reinforces the observation that the concept that is expressed earlier in the prompt has more chances to appear in the final image.

For SD, conditioned by CLIP text-encoders, the decreasing trend in the occurrence of objects as their position becomes more distant in the prompt may be partly due to the auto-regressive nature of the encoder. During self-attention, tokens only receive context from the elements to their left (beginning of the prompt). Tokens are thus devoid of the contexts of subsequent objects, while these last carry the context of the earlier words. Hence, the U-net model receives a more ambiguous signal from distant tokens.

The explanation is less clear with the T5 encoder, where all words have access to each other during self-attention. Used in IF, one can see that the third and fourth objects have the same chance to appear, but it is significantly less than the object in the second position, itself below the first one expressed in the prompt. We hypothesize that during cross-attention, the models learned to give more importance to tokens with earlier positions due to the training data, which typically places important elements related to the image at the beginning of the caption. Asserting this explanation would nevertheless require significant work to analyze the original training dataset used to pre-train the six generative models, which is out of the scope of this paper.

\begin{figure}
    \centering
    \includegraphics[width=0.8\linewidth]{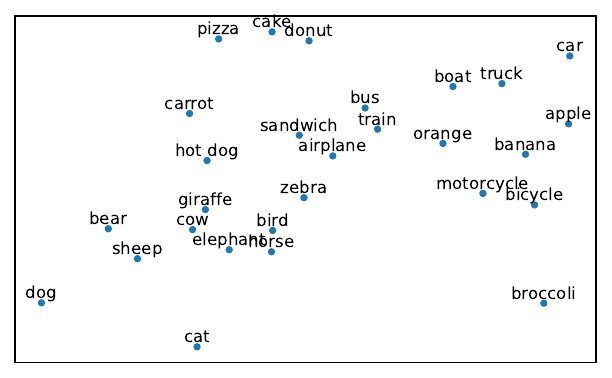}
    \caption{MDS on the objects score dissimilarity for SD 1.4.}
    \label{fig:mds_sd14}
\end{figure}
Finally, we investigated the impact of semantic relationships between objects within the prompt on the ability of the model to generate both of them. Considering a template with two objects, we hypothesized that the T2I model would fail more often to represent in the same image two objects that are semantically linked. We considered 28 COCO labels from three macro-classes (\textit{vehicles, animals}, and \textit{foods}) and generated images using a template with 2 objects. From the resulting TIAM scores, we derived a dissimilarity metric between all objects (see Supp. Mat.~\ref{appendix:SemanticLink}) and projected all the labels with  Multidimensional Scaling (MDS). The resulting projections  (Fig.~\ref{fig:mds_sd14} for SD 1.4)  can be interpreted such that the closer two labels are, the more challenging it becomes for the model to represent them together. Across all models, we observe labels from the same class being clustered together, particularly for \textit{animals} and \textit{vehicles}. This suggests the presence of semantic proximity. We measured the semantic distance between both named objects with various methods (such as Wu-Palmer, Cosine similarity of CLIP text embeddings, or even attention's key-value representation) for SD 1.4 and SD 2 but the correlation with the TIAM score was only slightly negative. This indicates that semantic linking has probably either a small or indirect link to the alignment performance, but in any case, further research is needed to clarify this point.

\subsection{Attribute binding\label{sec:4_4_attributeBinding}}

In this section, we characterize the capacity of models to apply attributes to objects. Let us consider 2 prompts with respectively one and two objects using the same set $\mathcal{O}=\{$\textit{car, refrigerator, giraffe, elephant, zebra}$\}$ and $\mathcal{A}$ the set of colors defined in Section~\ref{sec:attribute_definition}.

We start by computing TIAM regardless of the correctness of the attribute and report the results in Tab.~\ref{tab:colored_vs_no_colored}.
Without attributes (\ding{55}) the scores are close to those of Tab.~\ref{tab:benchmark_24} and adding attributes (\ding{51}) has a limited impact when one object only is requested. However, it significantly impacts the performance when two objects are present in the prompt.

\begin{figure}
    \centering
    \includegraphics[width = 0.8\linewidth]{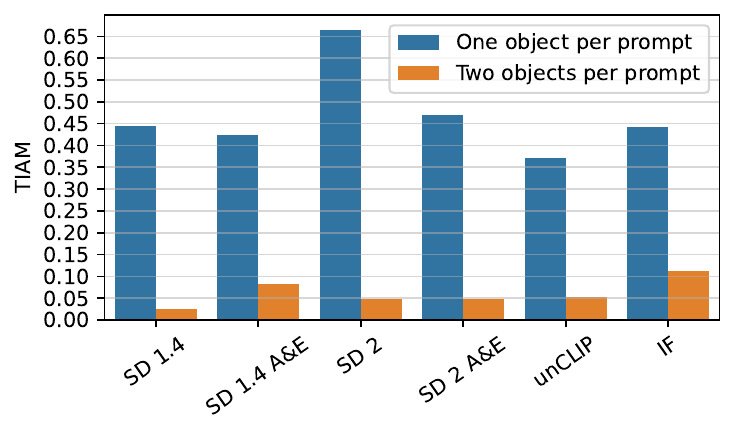}
    \caption{TIAM computed with object and color ground truth, for one and two colored objects per prompt.}
    \label{fig:tiam_colors}
\end{figure}

\begin{figure*}[tb]
    \centering
    \includegraphics[width=1\textwidth]{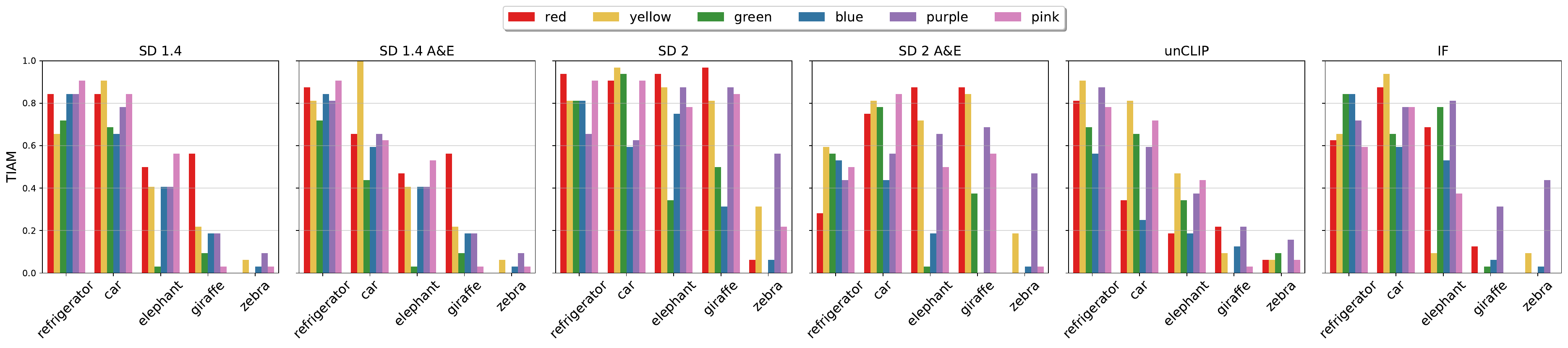}
    \caption{TIAM per color and object.}
    \label{fig:one_label_color_TIAM}
\end{figure*}

\begin{table}[t]
    \centering
    {\small{
            \begin{tabular}{lcccc}
                \toprule
                            & \multicolumn{2}{c}{1 object} & \multicolumn{2}{c}{2 objects}                         \\
                Attribute   & \ding{55}                    & \ding{51}                     & \ding{55} & \ding{51} \\
                \midrule
                SD 1.4      & 0.97                         & 0.95                          & 0.40      & 0.18      \\
                SD 1.4 A\&E & 0.93                         & 0.94                          & 0.64      & 0.42      \\
                SD 2        & 0.97                         & 0.97                          & 0.61      & 0.26      \\
                SD 2 A\&E   & 0.96                         & 0.96                          & 0.66      & 0.32      \\
                unCLIP      & 0.93                         & 0.89                          & 0.61      & 0.36      \\
                IF          & 0.98                         & 0.90                          & 0.61      & 0.49      \\
                \bottomrule
            \end{tabular}}}
    \caption{TIAM computed with object(s) ground truth only, with prompts containing attributes or not.
    }
    \label{tab:colored_vs_no_colored}
\end{table}

When TIAM is estimated by taking into account both the object and color ground truth, we observe that the models fail to assign the appropriate colors to the objects, even when these objects are present in the image (Fig.~\ref{fig:tiam_colors}). For instance, in the case of SD 1.4, objects are detected correctly in 95\% of single-object prompts, but only approximately 45\% of them have the requested colors.

In Fig.~\ref{fig:one_label_color_TIAM}, we report the attribution scores per color and object to analyze deeper the attributing ability of the models.
We see that models fail to generalize the binding to objects when it is uncommon to observe them in that particular color. As expected, it is indeed easier to assign colors to cars and refrigerators compared to animals. The models have likely been trained on photos of cars and refrigerators in various colors during training as it is more common to come across a green car than a green giraffe.

\begin{figure}
    \centering
    \includegraphics[width=0.8\linewidth]{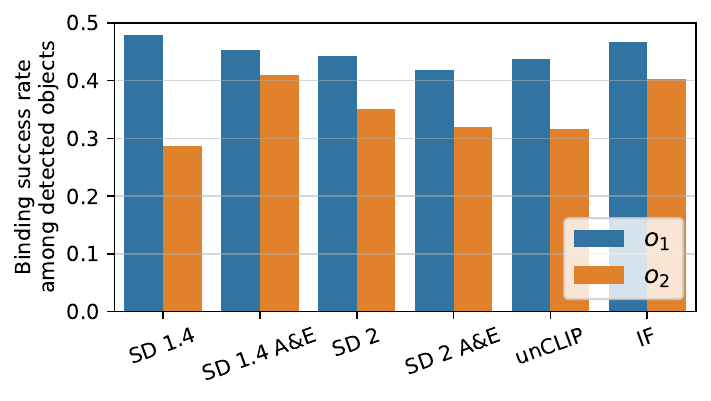}
    \caption{Success rate of color attribution w.r.t the detected objects.}
    \label{fig:binding_success_among_detected}

\end{figure}

Following the same approach as in Section~\ref{sec:catastrophic_neglect}, we also analyze the results concerning the position $i$ of attribute $a_i$ and object $o_i$ in the prompt involving 2 objects. In that case, the models succeed by a large margin to generate and bind the first object. However, knowing that the first object is more often generated, we compute a \textit{binding success rate}, which is the score of correctly attributed objects among the correctly generated objects (Fig.~\ref{fig:binding_success_among_detected}). The $o_2$ objects, however, continue to be less attributed. This reinforces the finding that the first object in the prompt has a greater influence on the final generation.
In the Supp. Mat.~\ref{appendix:attributebinding}, we report the results of the \textit{binding success rate} differentiated by colors for attributes in the first position and attributes in the second position.
We observed that the models face greater difficulty in assigning green and blue colors when two objects are involved (parallel with a single label case). It is worth noting that IF performs better than other models.

\subsection{Comparison to human, CLIP, BLIP}
We randomly chose 32 prompt-image pairs and asked 57 humans to assess the content alignment. Their reliability of agreement was 0.73 in terms of Fleiss'kappa~\cite{fleiss1971agreement}, which can be considered as ``Substantial''~\cite{landis1977agreement}. Half of the prompts had two objects and the 16 others had one colored object. The Pearson correlation of humans and TIAM was 0.82 ($p<10^{-8}$). We compared TIAM to two other automatic methods based on CLIP~\cite{radford2021learning} and BLIP~\cite{li2022blip}, which had respectively a correlation of 0.47 ($p<10^{-2}$) and 0.67 ($p<10^{-4}$) with humans. As detailed in the Supp. Mat.~\ref{appendix:compare_human_other}, TIAM had a better alignment with humans both for images with two objects and with one colored object.

\subsection{Toward noise mining?}\label{sec:illustrative_prompts}
To highlight the importance of noise performance, we selected seeds based on their TIAM score. In Fig.~\ref{fig:examples_wo_colors}, we present qualitative results using some of the worst and best seeds found with our approach, as well as prompts with objects and attributes that were not considered in our study.

Both with SD 2 and IF, the two objects are better represented in the image resulting from the ``good'' seed, showing that our method has the ability to find seeds that generalize to objects out of the domain of objects used to determine the best seeds.
We conducted the same seed selection process for both our worst and best seeds using colors in the prompt (Fig.~\ref{fig:examples_w_colors}). Once again the images resulting from the best seeds we identified with TIAM better reflect the prompt. With the ``bad'' seed, SD 2 suffers both from attribute leaking or binding (the ``blue moon'' is yellow, and the ``red lion'' is blue), and for both models, the images resulting from the ``good'' seed are more aligned. This emphasizes the dependency on the noise present at the beginning of the prompt, which remains a prominent factor of performance.
\newcommand{\wimage}{0.3}
\begin{figure}
    \centering
    \includegraphics[width=\wimage\linewidth]{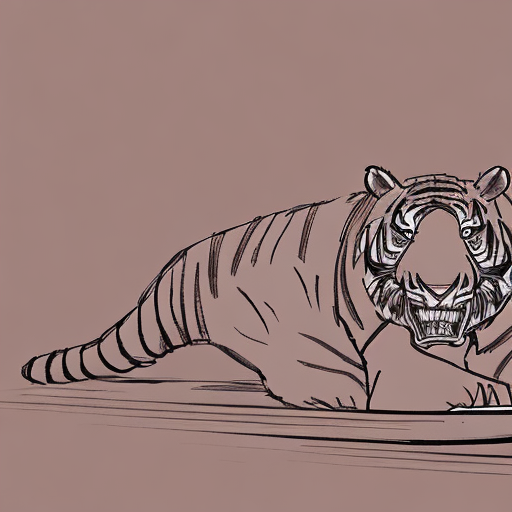}
    \hspace{-5px}
    \includegraphics[width=\wimage\linewidth]{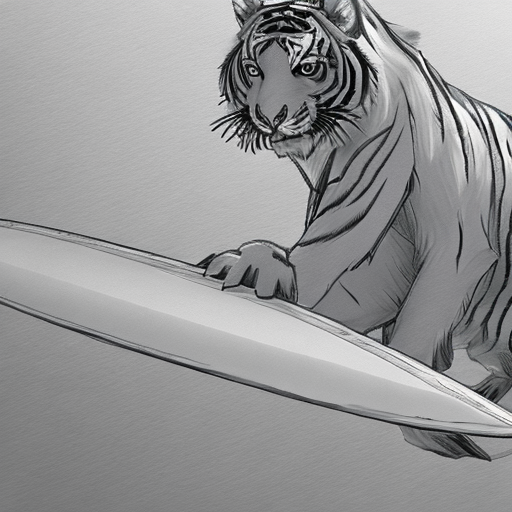}
    \\ \vspace{-1px}
    \includegraphics[width=\wimage\linewidth]{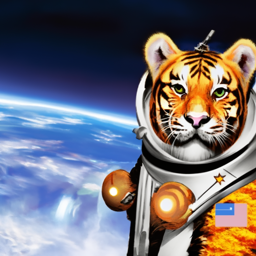}
    \hspace{-5px}
    \includegraphics[width=\wimage\linewidth]{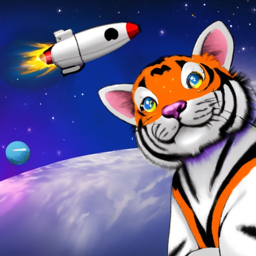}
    \caption{Generation of images with the same prompt and a ``bad'' (left) of a ``good'' (right) random seed. \textit{top}: SD 2 ``a sketch of a tiger and a surfboard, 4k, 8k, ghibli'' seed 23, 11. \textit{bottom}: IF ``a photo of space tiger and a rocket'' seed 41, 9.}
    \label{fig:examples_wo_colors}
\end{figure}

\begin{figure}
    \centering
    \includegraphics[width=\wimage\linewidth]{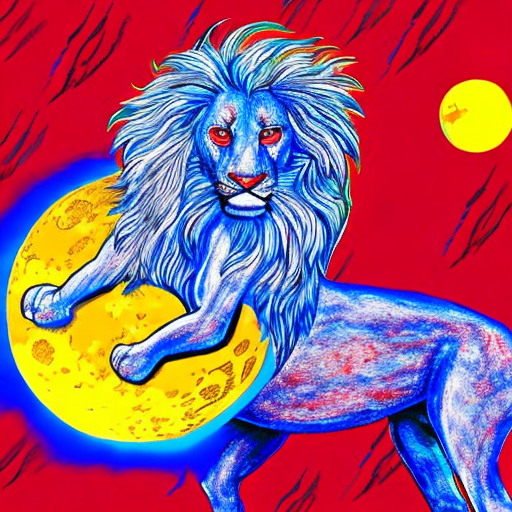}\hspace{-3px}
    \includegraphics[width=\wimage\linewidth]{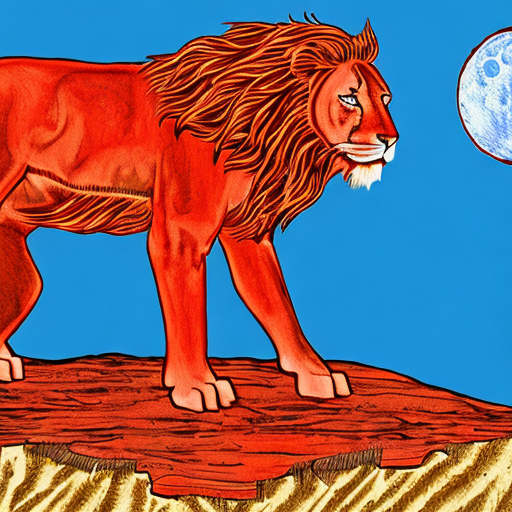}
    \\ \vspace{-1px}
    \includegraphics[width=\wimage\linewidth]{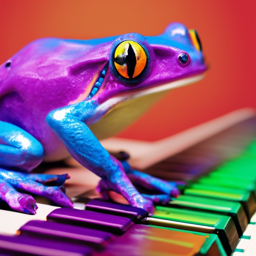}\hspace{-3px}
    \includegraphics[width=\wimage\linewidth]{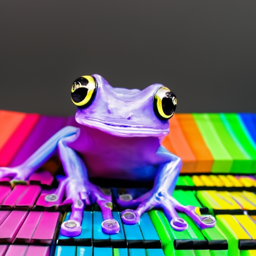}
    \caption{Generation of images with the same prompt and a ``bad'' (left) of a ``good'' (right) random seed. \textit{top}:
        SD 2 ``a drawing of a red lion and a blue moon, pop art, 4k, highly detailed'' seed 27, 17. \textit{IF} ``a photo of a purple frog and a rainbow piano'', seed 19, 24.}
    \label{fig:examples_w_colors}
\end{figure}

\section{Conclusion, Limits and Perspective}

We proposed a new metric to automatically quantify the performance of T2I models in terms of prompt-image alignment. Contrary to previous research efforts, it is based on prompt templates, that allow a finer analysis with regard to the syntax of the prompt. Hence, we showed that the alignment performance of most T2I models drops significantly with the number of objects specified in the prompt and that effect is even more critical for color attribution. Extending TIAM with more objects and attributes would result in the need to generate an exponential number of prompts, thus becoming cumbersome in practice. However, we show in Section~\ref{appendix:scalabilityTiam} of the Supp. Mat. that TIAM can be reliably estimated from a set of $\approx 300$ prompts.
Our metric also allows us to study the influence of the input seed at inference. We showed that there exist some seeds that systematically result in better output images than others and that it generalizes to objects out of the set used to determine them. It draws possible future research toward the mining of such ``good seeds'', similarly to some studies for text models \cite{dodge-arxiv-20}, as a complementary activity to prompt engineering to optimize the outputs of T2I models.
For a comprehensive analysis of T2I performance, TIAM should be combined with other metrics reflecting other aspects than prompt-image alignment, such as \cite{salimans2016improved_gans, heusel2017fid,cho2022dalleval,saharia2022photorealistic,zhang2023auditing}.

\paragraph{Acknowledgement} This work was granted access to the HPC resources of IDRIS under the allocation 2022-AD011014009 made by GENCI. This publication relied on the use of the FactoryIA supercomputer, financially supported by the Ile-de-France Regional Council.

    {\small
        \bibliographystyle{ieee_fullname}
        \bibliography{egbib}
    }

\clearpage
\appendix
\section*{Supplementary Material}

\newcommand{\mO}{\mathcal{O}}
\newcommand{\mA}{\mathcal{A}}
\newcommand{\mU}{\mathcal{U}}
\newcommand{\mP}{\mathcal{P}}
\newcommand{\mQ}{\mathcal{Q}}

\section{Proof of Proposition 1 and its Generalization \label{sec:proof_permutation}}
We give the proof of proposition 1, which is easy to establish, then we derive a formula giving the number of templates in a more general realistic case (proposition 3). For this, we introduce an intermediate step (proposition 2) and most importantly a set of notations that facilitates the derivation of the final analytic formula.

Let us consider a collection of object sets $\{\mathcal{O}_i\}_{i=1}^N$ and a collection of attribute sets $\{\mathcal{A}_i\}_{i=1}^N$ that are used to define a template with $N$ objects, each being qualified by an attribute. Hence, at position $i$ (without considering the context), the prompt results from the inference of the template and contains a named object $o_i\in\mathcal{O}_i$ that is qualified by a (color) adjective $a_i\in\mathcal{A}_i$.

\paragraph{Proposition 1:} if $|mO|\ge N$, $|\mA|\ge N$ and $\forall i\in [\![1,N]\!], \mA_i=\mA, \mathcal{O}_i=\mO$ and $\forall (i,j)\in [\![1,N]\!]^2$ s.t $i<j$, we force $a_i\ne a_j$ and $o_i\ne o_j$, thus the number of unique prompts generated by the template is $\frac{|\mO|!|\mA|!}{\left(|\mO|-N\right)!\left(|\mA|-N\right)!}$

\begin{proof}
    Without attribute, each prompt contains $N$ objects that should be different; thus the number of unique possible prompts is the number of (arrangements) $N$-permutations of $|\mO|$ thus $\frac{|\mO|!}{\left(|\mO|-N\right)!}$. Similarly, if one considers the attributes only that must be different at each position $i$, we have  $N$-permutations of $|\mA|$. Finally, since both are independent, the final number of unique prompts generated by the template is the product of both.
\end{proof}

We can consider a generalization where the sets are different at each position, that is we remove the conditions $a_i\ne a_j$ and $o_i\ne o_j$ from proposition 1. For example, such a template could be defined to have a vehicle at the first position, a fruit at the second position, and an animal at the third one. The intersection of the sets at each position may also be non-empty, for example, the vehicle may be $\{blue, red, green, black, yellow, white\}$, the fruit $\{red, green, yellow\}$, and the animal $\{green, black, yellow, white\}$. Similarly, the attributes can be repeated in the prompt \textit{e.g.} if one wants ``a yellow car and a red apple and a red elephant''.

\paragraph{Proposition 2:} if $\forall i \in [\![1,N]\!]$, $|\mO_i|\ge N$, $|\mA_i|\ge N$ thus the number of unique prompts generated by the template is $\prod_{i=1}^N |\mA_i|.|\mO_i|$. Hence, if  $\forall i\in [\![1,N]\!], \mA_i=\mA, \mO_i=\mO$, thus the number of unique prompts generated by the template is $(|\mO|.|\mA|)^N$

\begin{proof}
    At any position $i$ any all the attributes $a_i\in\mA_i$ can be associated with the object $o_i\in\mO_i$ thus it gives $|\mA_i|.|\mO_i|$ possibilities. Since repetitions are allowed, we have a structure of arborescence (rooted tree), thus the total number of prompts is the product at any position, thus $\prod_{i=1}^N |\mA_i|.|\mO_i|$.
\end{proof}

However, if the sets of objects are not exclusive we consider that we should not allow any repetition, since it would be strange to require ``a car and an apple and a car''. Such repetition can nevertheless be allowed if the object appears several times with different attributes. In other words, if $\forall i \in [\![1,N]\!]$, $|\mO_i|\ge N$, $|\mA_i|\ge N$  and $\forall (i,j)\in [\![1,N]\!]^2$ s.t $i<j$ we force $o_i\ne o_j$ if $a_i=a_j$.

\begin{table}[t]
    \centering
    \begin{tabular}{c c c}
        \hline
        $\mU_1$    & $\mU_2$       & $\mU_3$      \\
        \hline
        blue truck & green banana  & green tiger  \\
        red truck  & purple banana & blue tiger   \\
        blue bike  & green cherry  & green bear   \\
        red  bike  & purple cherry & blue bear    \\
        red car    & purple apple  & yellow apple \\
        \hdashline
        blue car   &               & blue car     \\
                   & green apple   & green apple  \\
        \hline
    \end{tabular}
    \caption{Example of \textit{objects with attributes}, \underline{above dashed line}: that can appear at a specific location and are thus in $\mU_i^0$; \underline{below dashed line}: that can appear at several locations in the prompt. One notes that in that case $\mO_1=\{'truck','bike','car'\}$ and $\mA_1=\{'blue','red'\}$}.
    \label{tab:ex_colored_objects}
\end{table}

\begin{table}[t]
    \centering
    \begin{tabular}{c c c| c}
        \hline
        \multicolumn{3}{c|}{configuration} & $\mP$                                 \\
        pos. 1                             & pos. 2      & pos. 3      &           \\
        \hline
        blue car                           & green apple & $\mU_3^0$   & $\{3\}$   \\
        blue car                           & $\mU_2^0$   & green apple & $\{2\}$   \\
        blue car                           & $\mU_2^0$   & $\mU_3^0$   & $\{2,3\}$ \\
        $\mU_1^0$                          & green apple & blue car    & $\{1\}$   \\
        $\mU_1^0$                          & green apple & $\mU_3^0$   & $\{1,3\}$ \\
        $\mU_1^0$                          & $\mU_2^0$   & blue car    & $\{1,2\}$ \\
        $\mU_1^0$                          & $\mU_2^0$   & green apple & $\{1,2\}$ \\
        \hline
    \end{tabular}
    \caption{Illustration of the set $\mP$ (fourth col.)  for the example of Tab.~\ref{tab:ex_colored_objects}. The notation $\mU_i^0$ means that any colored object of $\mU_i^0$ can be used at that position}
    \label{tab:definition_Q_P}
\end{table}

In that case, things depend on the overlap between the combination of attributes and objects at each position. In other words, if the attributes are colors, it depends on how many times the same colored object can appear at different positions in the prompt. Let us consider the sets $\mU_i=\mA_i\times\mO_i$ of \textit{objects with attributes} at position $i$ in the prompt. Since we have $o_i\ne o_j$ if $a_i=a_j$, each $\mU_i$ can be split into two disjoint subsets. The subset $\mU_i^0$ contains the objects with attributes that can appear at position $i$ only, and its complement in $\mU_i$ contains the objects with attributes that can appear at another position $j\ne i$ in the prompt.

The number of prompts that the template can generate depends on the  overlap between these complements of the $\mU_i^0$. For instance, the ``blue car'' can appear at positions 1 and 3, while the ``green apple'' can appear at positions 2 and 3 as illustrated in Tab.~\ref{tab:ex_colored_objects}. The objects that can appear at several positions define some configurations that are uniquely defined by the set of positions  where they appear in the prompt.

Symmetrically, these configurations are also defined by the sets of positions where only one object appears at a single position (thus from $\mU_i^0$).
We note $\mP$ the set of these last positions
(See Tab.~\ref{tab:definition_Q_P} for an illustration of these sets). Note that the same element can appear several times in $\mP$, 
actually when several objects with attributes can appear at the same position (\textit{e.g.} both ``blue car'' and ``green apple'' can appear at position 3 in our example). Formally, $\mP$ is a set that contains the sets of indexes of the non-empty positions of all the N-uples that verify  $(\mU_i \text{\textbackslash} \mU_i^0)\cap(\mU_j \text{\textbackslash} \mU_j^0)\ne\emptyset$.
Using this formalization of the problem, we have:
\paragraph{Proposition 3:} Let $\forall i \in [\![1,N]\!]$, $|\mO_i|\ge N$, $|\mA_i|\ge N$. If $\forall (i,j)\in [\![1,N]\!]^2$ s.t $i<j$ and $a_i=a_j$, we force $o_i\ne o_j$, thus the number of unique prompts generated by the template is: 
\begin{equation}\label{eq:prop1_generalized}
    \prod_{i=1}^N |\mU_i^0| +  \sum_{P\in\mP} \prod_{i\in P}|\mU_i^0|
\end{equation}

\begin{proof}
    Since the elements of the $\mU_i^0$ are different and appear at a unique position, they generate $\prod_{i=1}^N |\mU_i^0|$ templates.

    For any configuration in $\mP$, an object with attributes (from one of the $\mU_i\text{\textbackslash}\mU_i^0$) appears at most once since if $a_i=a_j$ thus $o_i\ne o_j$.

    For each configuration $P\in\mP$ (such as the lines of  Tab.~\ref{tab:definition_Q_P}) the number of templates generated is the product of all the $|\mU_i^0|$ for that configuration, thus $\prod_{i\in P}|\mU_i^0|$ for each configuration of position $P$. If we sum for all the possible $P$ in $\mP$, it results in the number of templates generated with objects with attributes that can appear at several positions.

    The sum of both terms gives the value in \autoref{eq:prop1_generalized}
\end{proof}
Hence the exact number of prompts generated depends on the possible overlaps between objects with attributes, both the number of elements and the place they can appear or not. By introducing more notations on these elements and their position, one could derive a formula but it would be tedious, without obvious interest in practice. Indeed, $\mP$ can be built easily through a tree structure, by considering iteratively all the sets $(\mU_i \text{\textbackslash} \mU_i^0)\cup\{\emptyset\}$, and allowing only the nodes that did not previously appear and the branch corresponding to a non-empty set in $\mP$ (such as in Tab.~\ref{tab:definition_Q_P}).

\section{Prompt Templates \label{appendix:prompt_template}}
We detail the templates and the COCO labels used in the study.

\subsection{Without attribute}
The 24 COCO labels used for the study of Sections~\ref{sec:preliminary} and \ref{sec:random_seed} are the following: \textit{
    bicycle,
    car,
    motorcycle,
    truck,
    fire hydrant,
    bench,
    bird,
    cat,
    dog,
    horse,
    sheep,
    cow,
    elephant,
    bear,
    zebra,
    giraffe,
    banana,
    apple,
    broccoli,
    carrot,
    chair,
    couch,
    oven,
    refrigerator}. The prompt template is ``a photo of $det(o_1)$  $o_1$ and  $det(o_2)$  $o_2$''.

For the semantic studies, end of Section~\ref{sec:catastrophic_neglect}, we use the vehicles, animals, and food labels \ie \textit{bicycle, car,
    motorcycle,
    airplane,
    bus,
    train,
    truck,
    boat,
    bird,
    cat,
    dog,
    horse,
    sheep,
    cow,
    elephant,
    bear,
    zebra,
    giraffe,
    banana,
    apple,
    sandwich,
    orange,
    broccoli,
    carrot,
    hot dog,
    pizza,
    donut,
    cake}.
The template used is ``a photo of $det(o_1)$  $o_1$ and  $det(o_2)$  $o_2$''.

To evaluate the capacity of models to represent multiple objects, we use the following template :
\begin{itemize}
    \item 1 object: ``a photo of $det(o_1)$  $o_1$''
    \item 2 objects: ``a photo of $det(o_1)$  $o_1$ and  $det(o_2)$  $o_2$''
    \item 3 objects: ``a photo of $det(o_1)$  $o_1$'' next to $det(o_2)$  $o_2$ and $det(o_3)$  $o_3$''
    \item 4 objects: ``a photo of $det(o_1)$  $o_1$'' next to $det(o_3)$  $o_2$ with $det(o_3)$  $o_3$ and $det(o_4)$  $o_4$''
\end{itemize}

We do not use commas because it tends to reduce the score. We report in Tab.~\ref{tab:TIAM_3_comma} the comparison between the prompt ``\textit{a photo of} $det(o_1)$  $o_1$ \textit{next to} $det(o_2)$  $o_2$ \textit{and} $det(o_3)$  $o_3$'' and the prompt with a comma ``\textit{a photo of} $det(o_1)$  $o_1$, $det(o_2)$ $o_2$ \textit{and} $det(o_3)$  $o_3$''.

\begin{table}[]
    {\small{\begin{center}
                \begin{tabular}{llll}
                    \toprule
                    Model       & w comma & w/o comma         \\
                    \midrule
                    IF          & 0.12    & 0.21 $_{(+0.09)}$ \\
                    SD 1.4      & 0.01    & 0.02 $_{(+0.01)}$ \\
                    SD 1.4 A\&E & 0.13    & 0.17 $_{(+0.04)}$ \\
                    SD 2        & 0.13    & 0.15 $_{(+0.02)}$ \\
                    SD 2 A\&E   & 0.17    & 0.21 $_{(+0.04)}$ \\
                    unCLIP      & 0.13    & 0.10 $_{(-0.03)}$ \\
                    \bottomrule
                \end{tabular}

            \end{center}}}

    \caption{Comparison of the TIAM for two templates with 3 entities, separated by commas or related with words.}
    \label{tab:TIAM_3_comma}
\end{table}

\subsection{With Attribute}
To evaluate the attribute binding in Section~\ref{sec:4_4_attributeBinding}, we used the following templates:
\begin{itemize}
    \item one object : ``a photo of $det(a_1)$ $a_1$ $o_1$''
    \item two objects : ``a photo of $det(a_1)$ $a_1$ $o_1$ and $det(a_2)$ $a_2$ $o_2$ ''
\end{itemize}
As reported in Sections~\ref{sec:attribute_definition} and \ref{sec:4_4_attributeBinding} of the main paper, we have $\mathcal{O}=\{$\textit{car, refrigerator, giraffe, elephant, zebra}$\}$ and $\mathcal{A}=\{\mathit{red, green, blue, purple, pink, yellow}\}$.

\section{Reference Colors and Other Possible Attributes (Size, Texture)}\label{appendix:other_attributes}
We plot in Fig.~\ref{fig:best_example} the interpolation of the best examples in Lab on the CIE 1931 Chromaticity Diagram. The exact chroma values were extracted from Fig.1 in \cite{kay2016color_survey}, available at \href{https://doi.org/10.1007/978-1-4419-8071-7_113}{this link}.
\begin{figure}
    \begin{center}
        \includegraphics[width=0.8\linewidth]{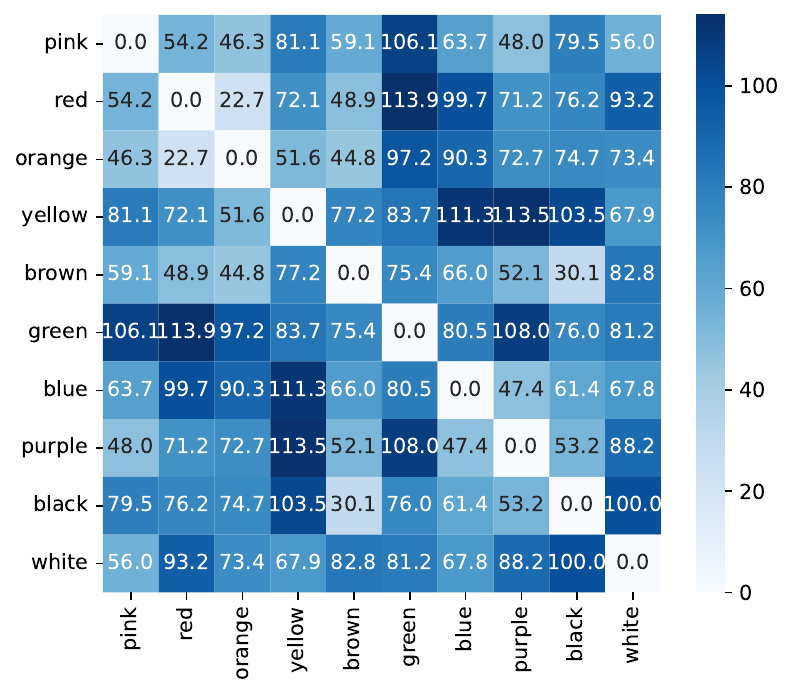}
        \caption{L2 norm distance between our reference colors.}
        \label{fig:distance_colors}
    \end{center}
\end{figure}
We do not use our reference colors  \textit{orange} and \textit{brown} because they are too close to \textit{red} and \textit{black} respectively in the CIELab space (Fig.~\ref{fig:distance_colors}).

\begin{figure}
    \begin{center}
        \includegraphics[width=0.8\linewidth]{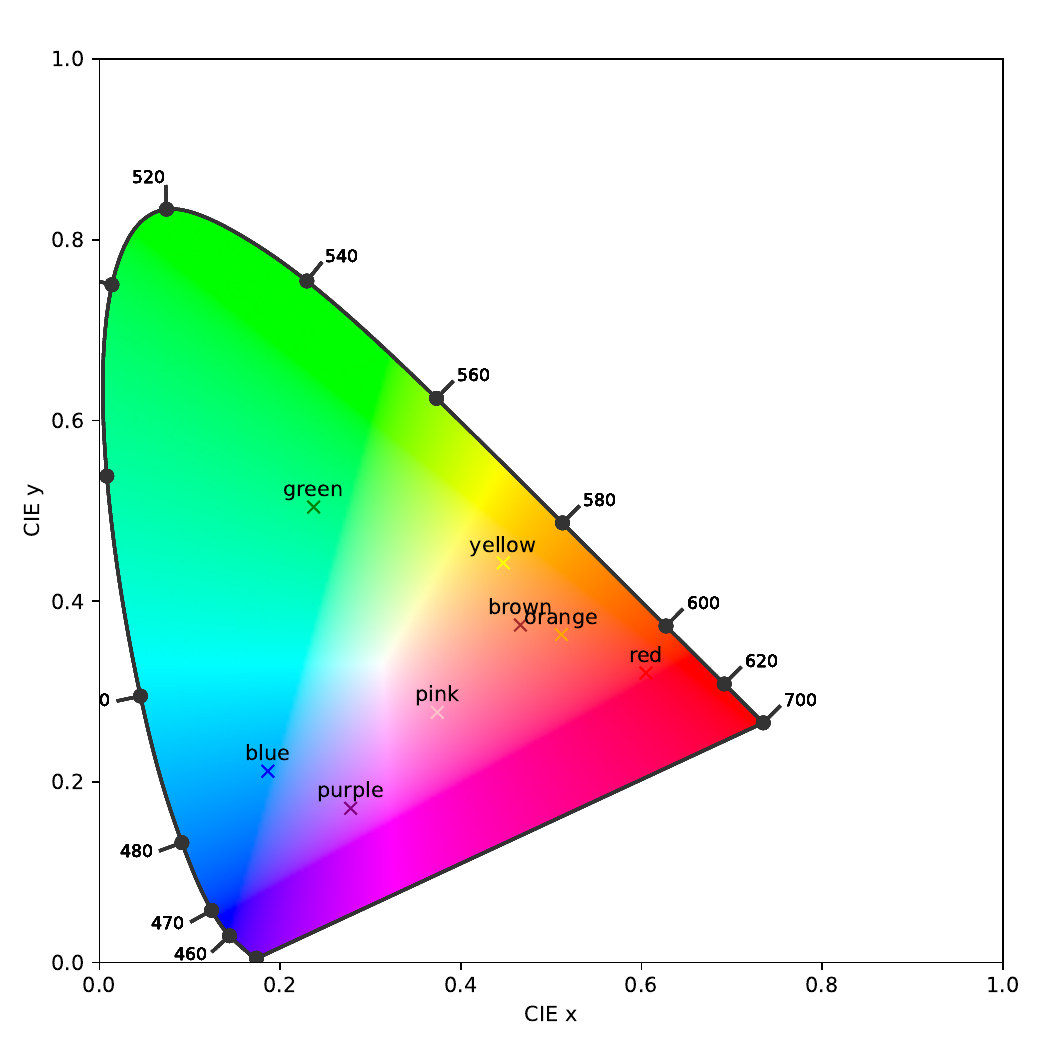}
        \caption{CIE 1931 Chromaticity Diagram with the best example for each color. When multiple best examples for one color, we compute one best example by averaging the value in the CIELAB space.
        }
        \label{fig:best_example}
    \end{center}
\end{figure}

To consider an attribute in TIAM, two crucial points need to be considered:
\begin{itemize}
    \item being able to extract the attribute from the image, with a sufficient level of reliability
    \item being able to name the attributes with unambiguous words in the prompt, ideally in several languages
\end{itemize}
The attribute \textit{colors} have the advantage of quite easily meeting these two conditions, as explained in Section~\ref{sec:attribute_definition} of the main paper, in particular, thanks to the works of Berlin and Kay \cite{BerlinBasicColor1969}. In other cases, however, this may prove trickier.

To extract the attribute \textit{size}, one can rely on the bounding box of the object detector. However, to determine whether an object is \textit{large}, \textit{medium}, or \textit{small} for example, it may also require to estimate its depth in the image (that can be done from monocular images to a certain extent~\cite{saxena2023monocular}), possibly an estimation of the intrinsic parameter matrix of the camera (while the image is generated) as well as the knowledge of the typical dimensions of the considered object or living being. Each of these estimations is a potential source of approximation that challenges the reliability of the final judgment.

Naming the \textit{size} may also be diverse. In the example above, \textit{large} may be replaced by \textit{big} in English and \textit{small} by \textit{tiny} in some cases. One could imagine relying on lists of synonyms or setting a threshold on the similarity to the textual embedding (\eg BERT) of an arbitrary predefined list of possible sizes (\eg large, medium, small), but there is no guarantee to get a list as unambiguous as in the case of colors. To our knowledge, there is no equivalent of the study of Berlin and Kay in that case. Moreover, references to sizes tend to include intensifiers more frequently than references to colors, like in expressions such as very small or fairly large, which adds an element of diversity. Finally, the way sizes are expressed can depend on forms of collocations. For instance, we can refer to a ``tall man'' but not to a ``tall balloon''. While this example can be probably explained by the form of the object, the easiest way to deal with such problem would be to collect co-occurrences from a corpus for (size adjective, noun) pairs and to use these co-occurrences as a filter after the generation of a prompt.

Extracting texture is a long-term and well-known task in computer vision~\cite{julesz1962visual_pattern} and many methods have been proposed to address it~\cite{hummeau2019texture_extraction_survey}. 
The question of naming the texture with unambiguous names may seem delicate. In the famous Brodatz dataset for example (available \href{http://sipi.usc.edu/database/database.php?volume=textures}{here}), one can note that several of them are named \textit{Wood shingle roof} (\href{https://sipi.usc.edu/database/database.php?volume=textures&image=41#top}{here} and \href{https://sipi.usc.edu/database/database.php?volume=textures&image=42#top}{here}), \textit{Brick wall} (\href{https://sipi.usc.edu/database/database.php?volume=textures&image=40#top}{here}, \href{https://sipi.usc.edu/database/database.php?volume=textures&image=47#top}{here}, or \href{https://sipi.usc.edu/database/database.php?volume=textures&image=38#top}{here}) or \textit{Sand} (\href{https://sipi.usc.edu/database/database.php?volume=textures&image=50#top}{here}, \href{https://sipi.usc.edu/database/database.php?volume=textures&image=51#top}{here} and \href{https://sipi.usc.edu/database/database.php?volume=textures&image=55#top}{here}) among others. Bhushan \textit{et al.} (1997) nevertheless identified a list of 98 representative words used to describe texture in English~\cite{bhushan1997texture_lexicon}
that was further reduced to 47 in the Describable Textures Dataset~\cite{compoi2014describable_texture}. However, such a number remains quite large (much more than the 11 colors of Berlin and Kay), such that the correspondence in other language than English is hazardous, not to mention the fact that some of them may seem ambiguous to several human users (our human study in Section~\ref{appendix:compare_human_other} includes users less than 10 years old with a limited vocabulary, as well as a majority of persons that are not computer vision scientists). Using Multidimensional Scaling (MDS) on these 98 words, Bhushan \textit{et al.} (1997) nevertheless identified 11 clusters that could be used, although naming these clusters is still problematic in practice. Finally, the easier approach may be to use three axes, identified with the MDS as well, namely:
\begin{itemize}
    \item repetitive versus nonrepetitive textures
    \item the nature of orientation: linearly oriented textures $\to$ multiple or no orientation $\to$ circularly oriented textures.
    \item complexity or simplicity of the surface
\end{itemize}

Hence, considering other attributes than \textit{colors} in TIAM is likely feasible, but integrating them neatly into the metric may require some work.

\section{Occurrence of Objects on Images \label{appendix:occurenceOfObject}}
\begin{figure}[h]
    \begin{center}
        \includegraphics[width=0.8\linewidth]{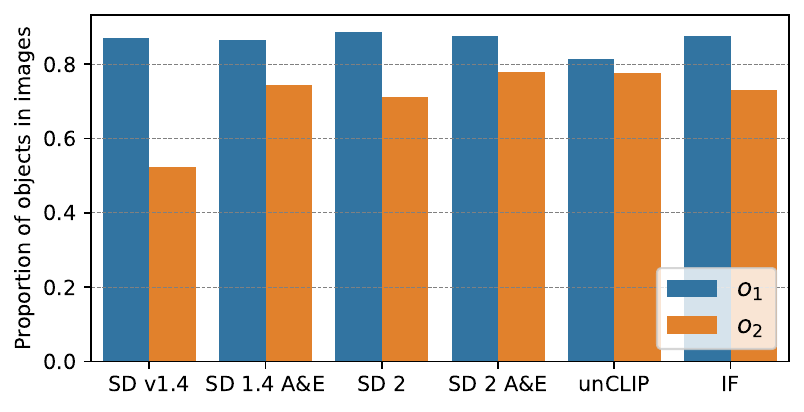}
        \caption{The proportion of occurrences of each object, based on its position in the prompt. The template of the prompt includes two objects.}
        \label{fig:apparition_2}
    \end{center}
\end{figure}

\begin{figure}[h]
    \begin{center}
        \includegraphics[width=0.8\linewidth]{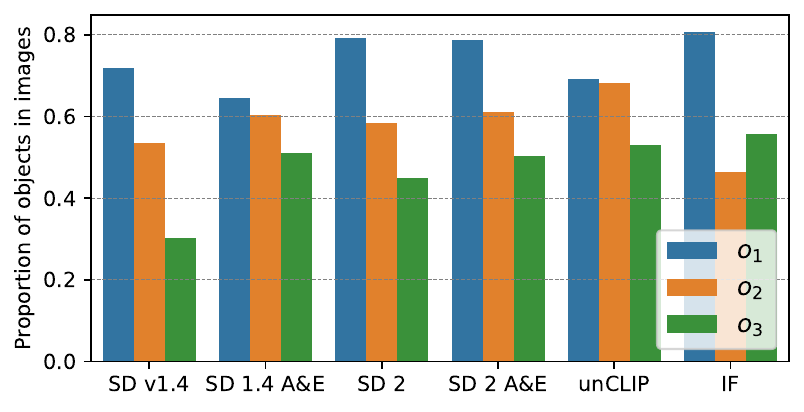}
        \caption{The proportion of occurrences of each object, based on its position in the prompt. The template of the prompt includes three objects.}
        \label{fig:appparition_3}
    \end{center}
\end{figure}

We show in our experiments (Section~\ref{sec:catastrophic_neglect}) that the initial objects in the template tend to appear more frequently than objects inserted subsequently and reinforce the observation that the concept that is expressed earlier in the prompt has more chances to appear in the final image. We present the result for two objects in Fig.~\ref{fig:apparition_2} and three objects in Fig.~\ref{fig:appparition_3}.

\section{Determining the Minimum Number of Images to Generate \label{appendix:minImgGenerate}}
\begin{figure}[h]
    \begin{center}
        \includegraphics[width=1.0\linewidth]{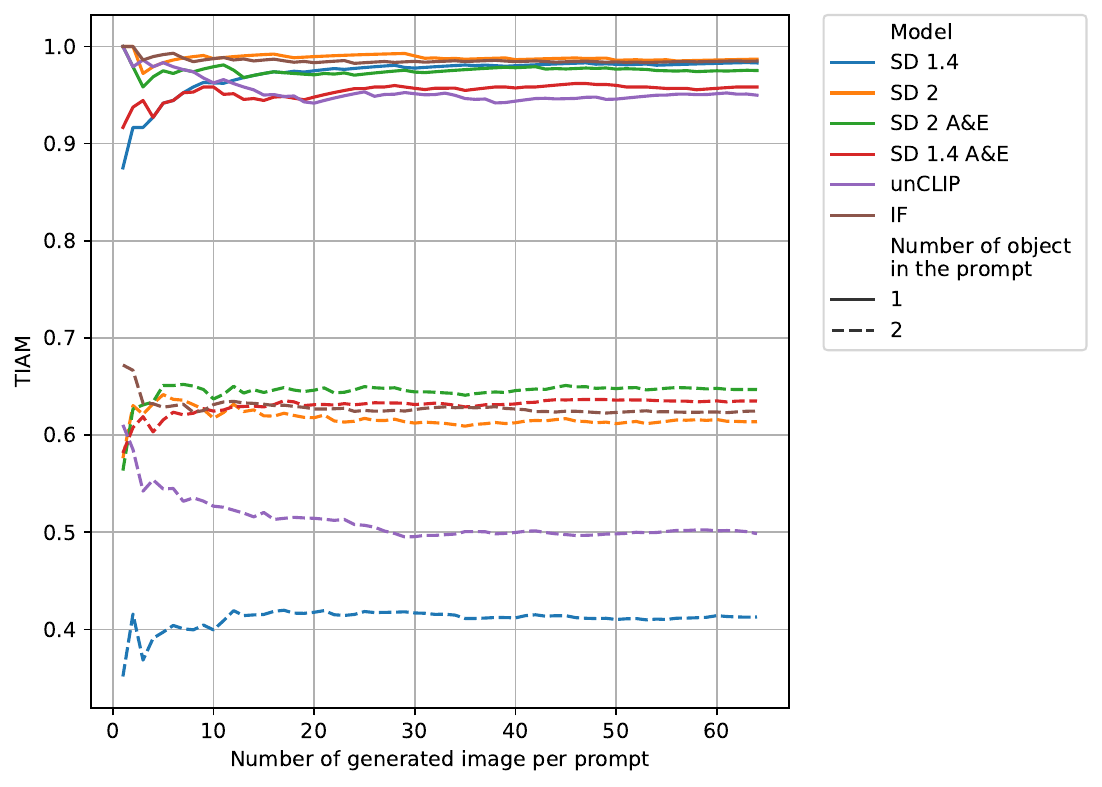}
    \end{center}
    \caption{TIAM as the function of the number of generated images per prompt.}
    \label{fig:scaling_one_label}
\end{figure}

In Fig.~\ref{fig:scaling_one_label} we report the TIAM score as a function of the number of generated images per prompt. The score stabilizes from 16 images. We chose to compute with 32 images to ensure robustness.

\section{Detection/Segmentation Details\label{appendix:detailSegmentationDetection}}
\begin{figure}[h]
    \begin{center}
        \includegraphics[width=0.8\linewidth]{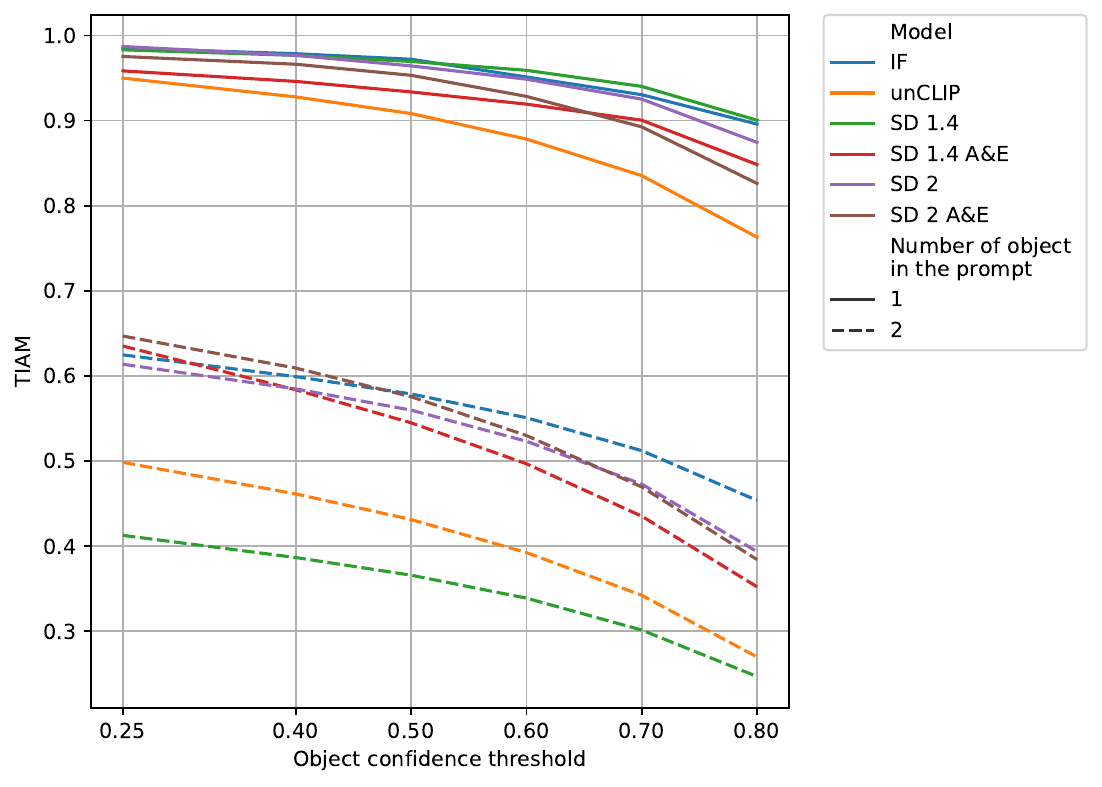}
        \caption{TIAM as a function of the YOLO object confidence threshold.}
        \label{fig:tiam_conf}
    \end{center}
\end{figure}

We use the largest YOLOv8 for segmentation\footnote{\href{https://github.com/ultralytics/assets/releases/download/v0.0.0/yolov8x-seg.pt}{https://github.com/ultralytics/assets/releases/download/v0.0.0/yolov8x-seg.pt}}. During segmentation inference, we set up the object confidence threshold for detection to 0.25 and the intersection over union IoU threshold for NMS to 0.8.
We compute the \textit{score} with different confidence thresholds and observe that the score decreases linearly as the threshold values increase (Fig.~\ref{fig:tiam_conf}).

\section{Text-to-Image Models Setup \label{appendix:setupModelT2I}}
All generations of images were done on Nvidia A100 SXM4 80 Go using \texttt{float16}. We list the main parameters for the different models used. If not mentioned, we use the default parameters from the library diffusers~\cite{von-platen-etal-2022-diffusers} (version 0.16.1).

\paragraph[SD 1.4 and SD 2]{
    SD 1.4 \footnote{\href{https://huggingface.co/CompVis/stable-diffusion-v1-4}{https://huggingface.co/CompVis/stable-diffusion-v1-4}} and SD 2 \footnote{\href{https://huggingface.co/stabilityai/stable-diffusion-2-base}{https://huggingface.co/stabilityai/stable-diffusion-2-base}}}
The models produce images of size 512$\times$512.

\begin{itemize}
    \item Guidance scale: 7.5
    \item Scheduler : DPMSolverMultistepScheduler \footnote{\href{https://huggingface.co/docs/diffusers/api/schedulers/multistep_dpm_solver}{https://huggingface.co/docs/diffusers/api/schedulers/multistep\_dpm\_solver}} \cite{lu2022dpmsolver}
    \item 50 inference steps
\end{itemize}

\paragraph[unCLIP]{unCLIP \footnote{\href{https://huggingface.co/kakaobrain/karlo-v1-alpha}{https://huggingface.co/kakaobrain/karlo-v1-alpha}}}
The model produces images of size 256$\times$256.
\begin{itemize}
    \item  25 inference steps for the prior, 25 inference steps for the decoder, and 7 steps of Super-resolution
    \item Prior guidance scale: 4, decoder guidance scale: 8
    \item Scheduler is the UnCLIPScheduler, a modified DDPM scheduler designed for this model.
\end{itemize}

\paragraph{IF}
It exists a different configuration of the Deepfloyd IF. We use for the first stage the L version\footnote{\href{https://huggingface.co/DeepFloyd/IF-I-L-v1.0}{https://huggingface.co/DeepFloyd/IF-I-L-v1.0}} with 100 inference steps and for the second stage the M version\footnote{\href{https://huggingface.co/DeepFloyd/IF-II-M-v1.0}{https://huggingface.co/DeepFloyd/IF-II-M-v1.0}} with 50 inference steps. We use both the DDPMS scheduler and guidance scale of 7 for the first stage and 4 for the second stage. With this configuration, we produce images of size 256$\times$256.

\section{Semantic Link\label{appendix:SemanticLink}}
\begin{figure}
    \centering
    \includegraphics[width=0.6\linewidth]{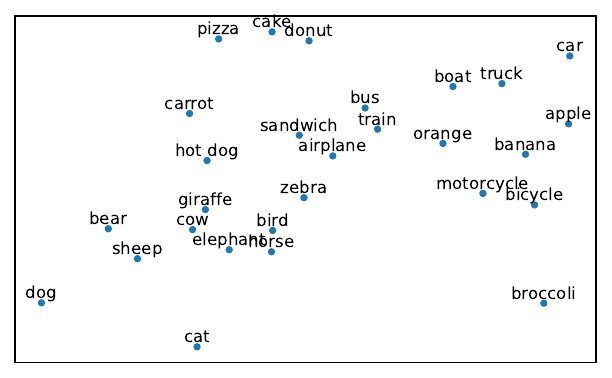}
    \caption{MDS on the objects score dissimilarity for SD 1.4.}
    \label{fig:mds_sd14_b}
\end{figure}

\begin{figure}
    \begin{center}
        \includegraphics[width=0.6\linewidth]{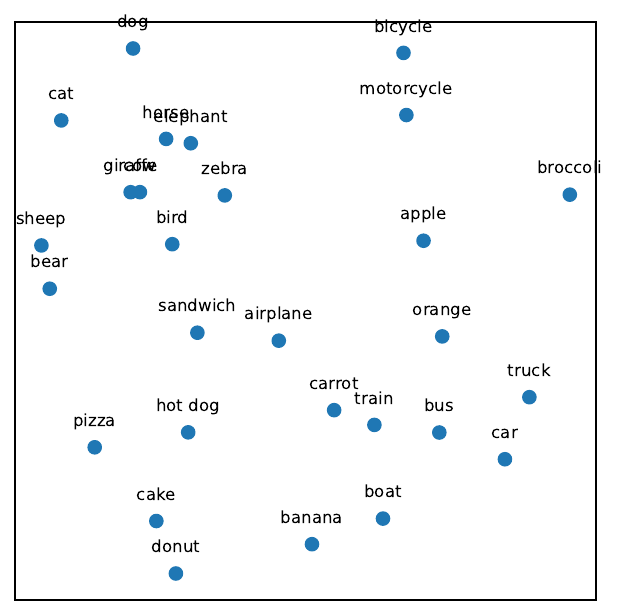}
        \caption{MDS on the objects score dissimilarity for SD 2.}
        \label{fig:mds_sd2}
    \end{center}
\end{figure}

\begin{figure}
    \begin{center}
        \includegraphics[width=0.6\linewidth]{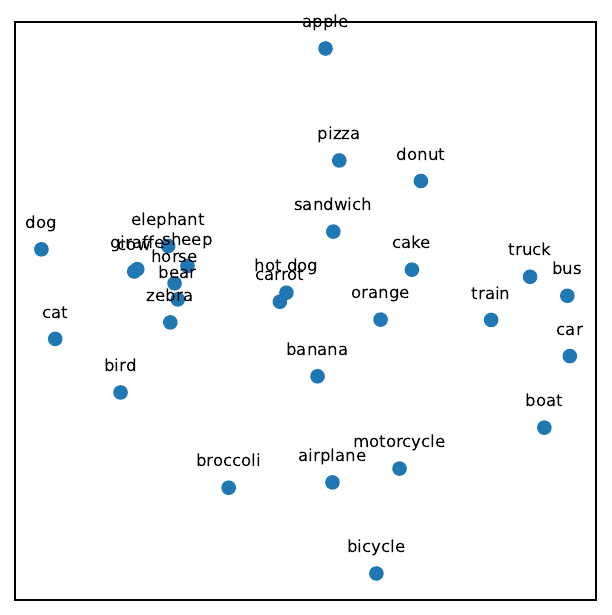}
        \caption{MDS on the objects score dissimilarity for unCLIP.}
        \label{fig:mds_unclip}
    \end{center}
\end{figure}

\begin{figure}
    \begin{center}
        \includegraphics[width=0.6\linewidth]{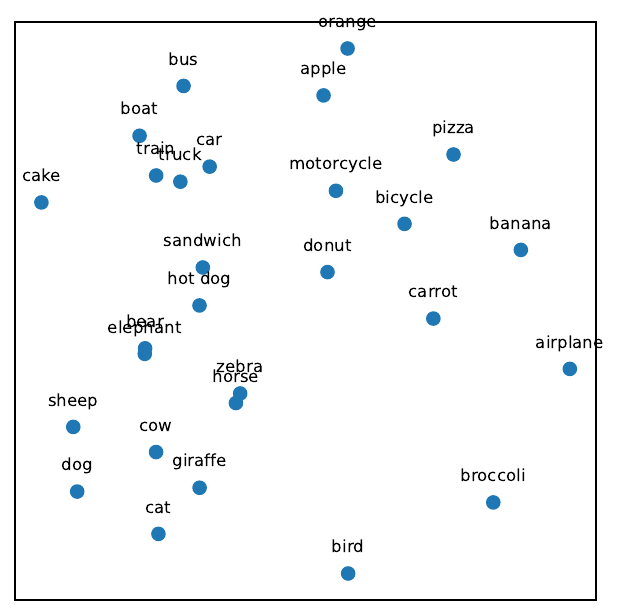}
        \caption{MDS on the objects score dissimilarity for IF.}
        \label{fig:mds_if}
    \end{center}
\end{figure}

To investigate the impact of semantic relationships between objects, we select 28 COCO labels from three macro-classes, \textit{vehicles, animals, and foods}, and generate images using a template with 2 objects. In order to study the influence of semantic links between objects with the same prompt, we consider the following dissimilarity metric on the set $\mO$.
\begin{multline}
    \forall o_x, o_y \in \mO, o_x \neq o_y, \\
    \hspace{1cm} d(o_x, o_y) = d(o_y, o_x) = \frac{\text{TIAM}_{z_i} + \text{TIAM}_{z_j}}{2}
\end{multline}
\begin{equation}
    \hspace{-3.2cm}\forall o_x \in \mO,\hspace{1cm} d(o_x, o_x) = 0
\end{equation}
where $z_i$ is $(o_x, o_y)$ and $z_j$ is $(o_y,o_x)$. For TIAM$_z$ we compute the score per $z$ (\ie per prompt). Using this dissimilarity between the labels, we project them with Multidimensional Scaling (MDS) into a 2D space, as represented in Fig.~\ref{fig:mds_sd14_b} for SD 1.4, in Fig.~\ref{fig:mds_sd2} for SD2, in Fig.~\ref{fig:mds_unclip} for unCLIP and in Fig.~\ref{fig:mds_if} for IF. The projection can be interpreted such that the closer two labels are, the more challenging it becomes for the model to represent them together. Note that we obtained similar projections with t-SNE instead of MDS.

For all the models, we observe some clusters of objects from the same macro-class, in particular \textit{animals}. It shows that \textbf{when two objects are semantically close, they tend to be harder being generated in the same image}. Tang \etal \cite{tang2022daam} obtain results in the same vein, showing that it is easier to generate a non-cohyponym than a cohyponym.

However, the effect remains slight, suggesting that the cohyponymy has either an indirect or minor link to this difficulty of generation. We quantified the effect by computing the correlation between the TIAM score for all the templates with two objects and the semantic distance between the two objects. We used various methods to estimate the semantic distance, including Wu-Palmer, the CLIPscore, and the cosine similarity between the embedding of the token in the prompt (before the attention of the transformer) for SD 1.4 and SD 2. For all distances, these correlations were negative (confirming the effect) but their absolute values were less than 0.5 (confirming the effect is slight).

\section{Attribute Binding\label{appendix:attributebinding}}

\begin{figure}
    \centering
    \includegraphics[width=0.8\linewidth]{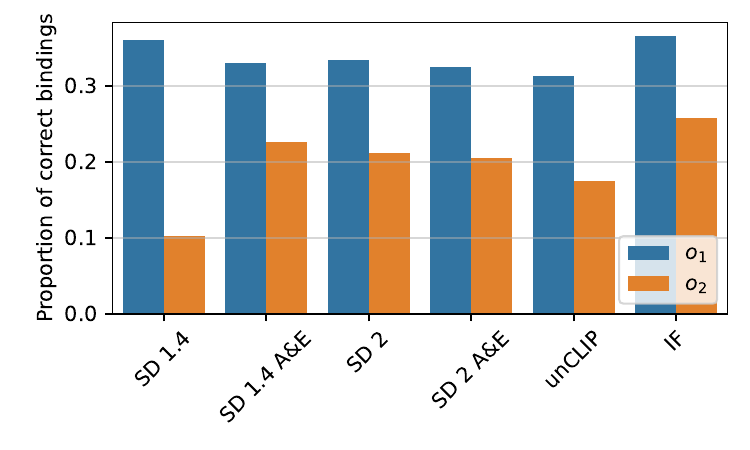}
    \caption{TIAM per object \ie proportion of correct generated object with the correct binding.}
    \label{fig:prop_detect_and_bind}
\end{figure}
\begin{figure}[h]
    \centering
    \includegraphics[width=0.8\linewidth]{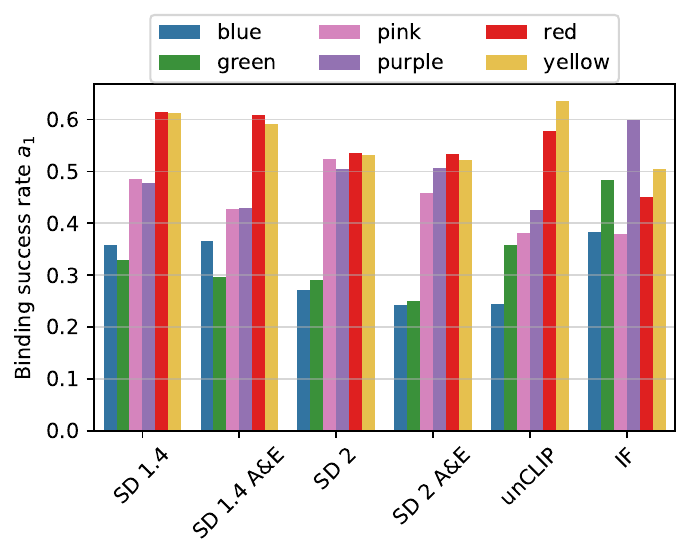}
    \caption{Binding success rate for the first object, among the first objects correctly detected.}
    \label{fig:binding_success_entity_1_color_among}
\end{figure}
\begin{figure}[h]
    \centering
    \includegraphics[width=0.8\linewidth]{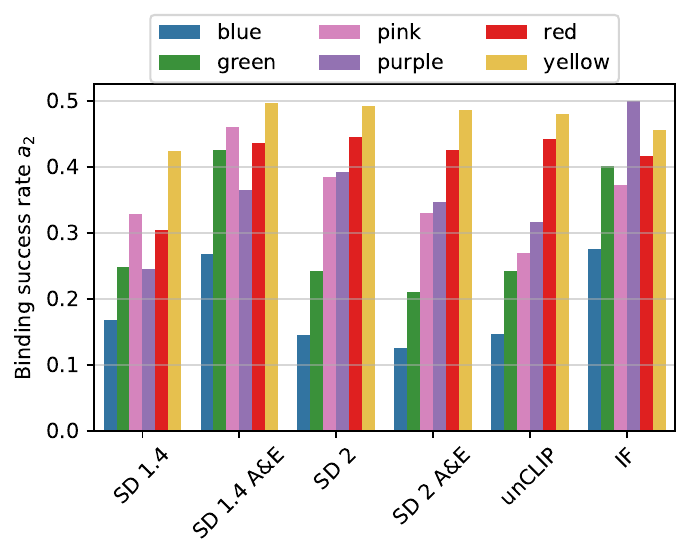}
    \caption{Binding success rate for the second object, among the second objects correctly detected.}
    \label{fig:binding_success_entity_2_color_among}
\end{figure}

We report the TIAM per object in Fig.~\ref{fig:prop_detect_and_bind}, showing that the first object is more often generated and correctly colored.
We compute the \textit{binding success rate}, but by differentiating by colors for attributes in the first position (Fig.~\ref{fig:binding_success_entity_1_color_among}) and attributes in the second position (Fig.~\ref{fig:binding_success_entity_2_color_among}). We observed that the models face greater difficulty in assigning green and blue colors when two objects are involved (parallel with a single label case). It is worth noting that IF performs better than other models.

\section{Latent Diffusion Model} \label{appendix:random_seed}
\begin{figure*}[]
    \begin{center}
        \includegraphics[width=0.8\linewidth]{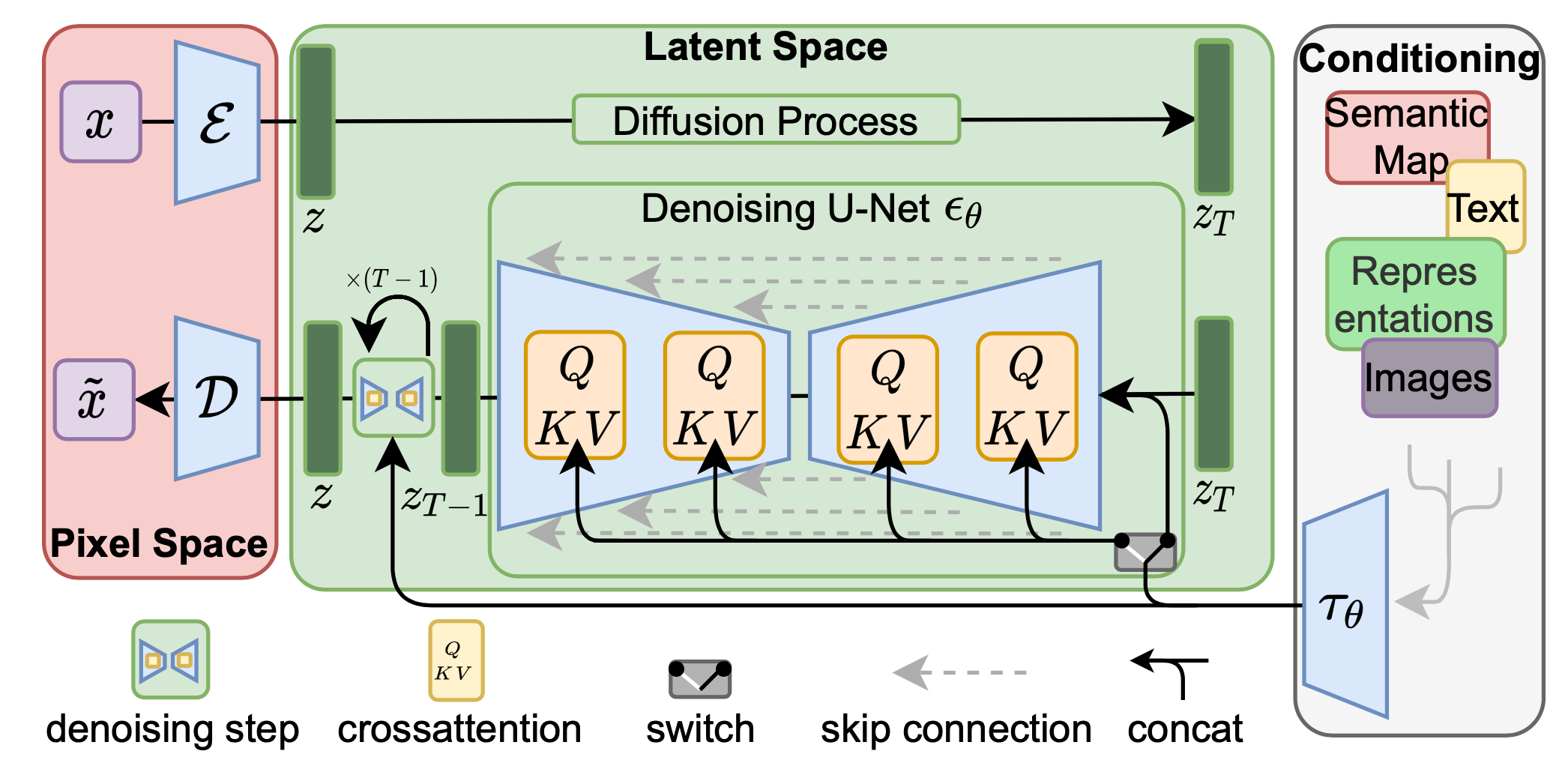}
    \end{center}
    \caption{Architecture of the Latent Diffusion Model \cite{rombach2021highresolution}.}
    \label{fig:archi_LDM}
\end{figure*}

We remind the architecture of the Latent Diffusion Model of Rombach \etal. (2021) in Fig.~\ref{fig:archi_LDM}, since the theoretical explanation of Section~\ref{sec:random_seed} of the main paper relies on it.

\begin{figure*}[h]
    \centering
    \includegraphics[width=1\linewidth]{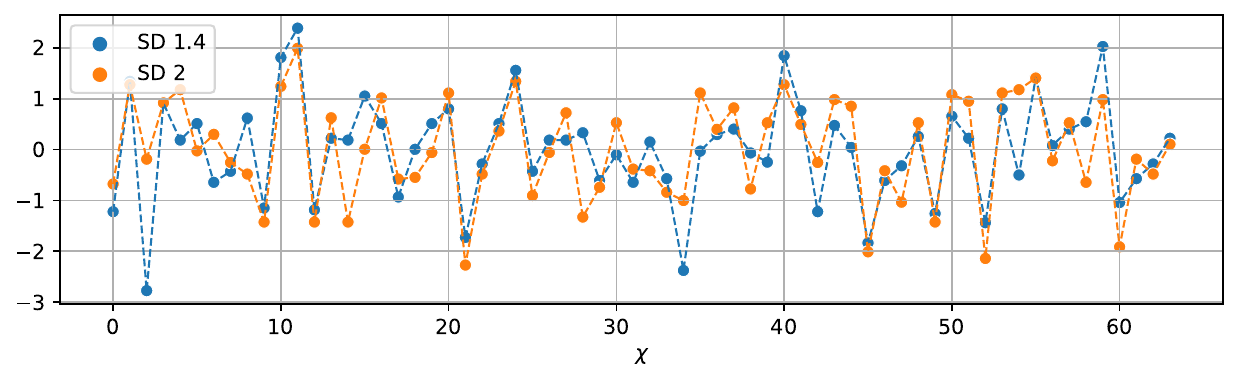}
    \caption{Standardized score per seed for SD 1.4 and SD 2. We observe that both of them have globally the same ``good'' and ``bad'' seeds (see Section \ref{appendix:random_seed} for explanations).}
    \label{fig:comparaison_seed}
\end{figure*}

During the investigation of the seed performances of the models, we made a noteworthy finding. We observed that the SD models exhibited similar behavior in relation to the seeds (if we do not consider the performance gap between the score of the model) \ie when we standardize the score of each seed for each model (Fig.~\ref{fig:comparaison_seed}) the score exhibits a remarkably similar trend for both models with the same \textit{``good''} and ``\textit{bad}'' seeds. However, we explain in the article (Section~\ref{sec:random_seed}) that the \textit{``good''} and ``\textit{bad}'' seeds are specific to each model, which may seem contradictory at first glance.

To explain this point, we need to remember the training of diffusion models. Let $x_0$ be an original image. We had on the image a scaled quantity of noise $\epsilon \sim \mathcal{N}(0,1)$ to obtain $x_t$ ($\mathbf{x}_t = \sqrt{\bar{\alpha}_t}\mathbf{x}_0 + \sqrt{1 - \bar{\alpha}_t}\boldsymbol{\epsilon}$) and the U-net $\epsilon_\theta$ try to predict with the added noise (the loss function ($\mathbb{E}_{\mathcal{E}(I),\epsilon\sim\mathcal{N}(0,1)}\left[ || \epsilon -\epsilon_\theta(x_t,t)||_2^2\right]$). At each training step a $t$ is drawn the model must predict the noise. In the case of LDM, just replace the $x$ with $z$ because the diffusion process is in the latent space.

\begin{figure*}[htbp]
    \centering
    \begin{tabular}{cc}
        \includegraphics[width=0.3\textwidth]{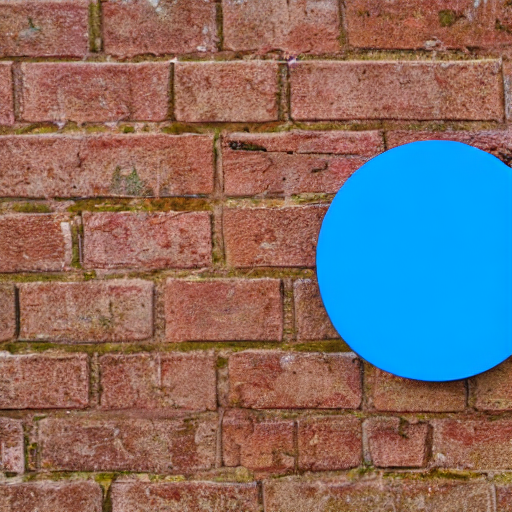}   &
        \includegraphics[width=0.3\textwidth]{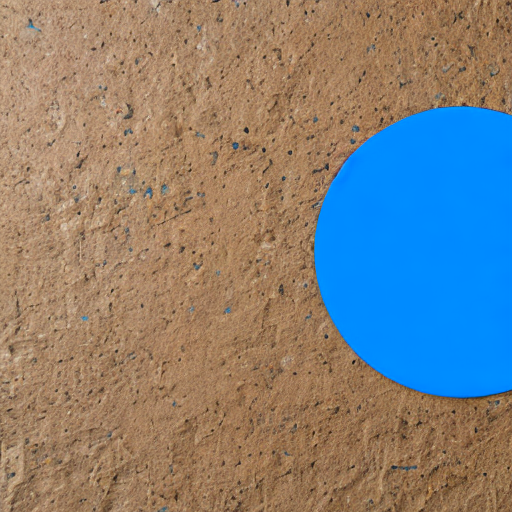}       \\
        \includegraphics[width=0.3\textwidth]{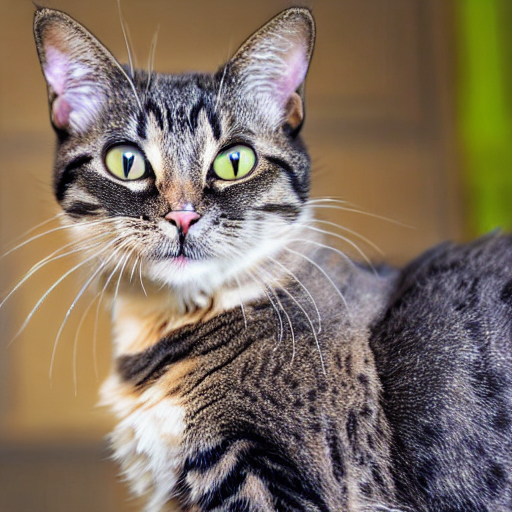}           &
        \includegraphics[width=0.3\textwidth]{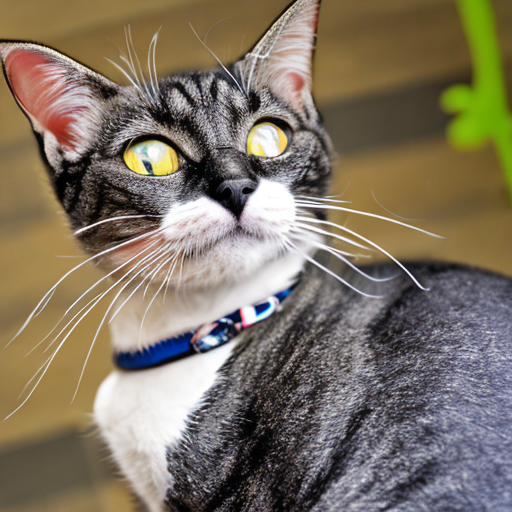}               \\
        \includegraphics[width=0.3\textwidth]{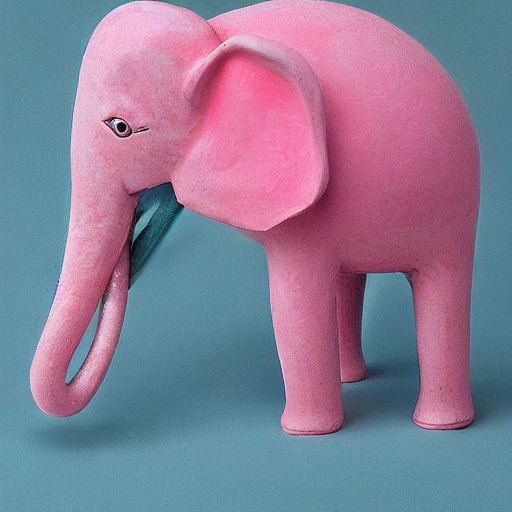} &
        \includegraphics[width=0.3\textwidth]{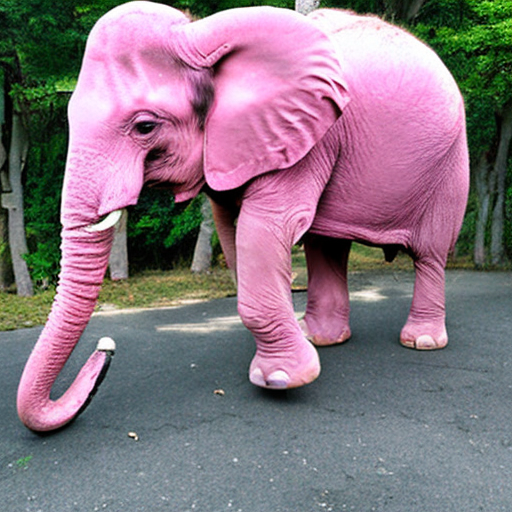}     \\
        \includegraphics[width=0.3\textwidth]{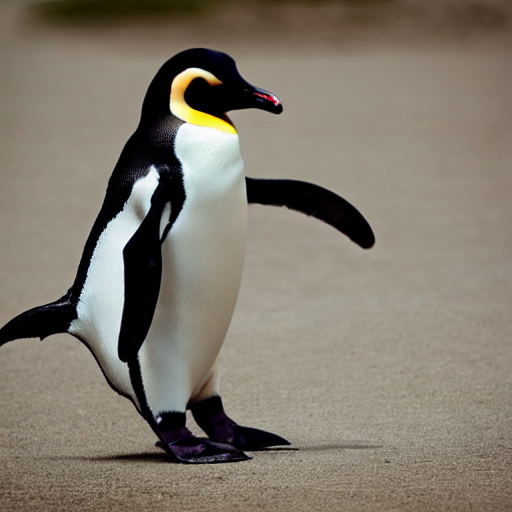}       &
        \includegraphics[width=0.3\textwidth]{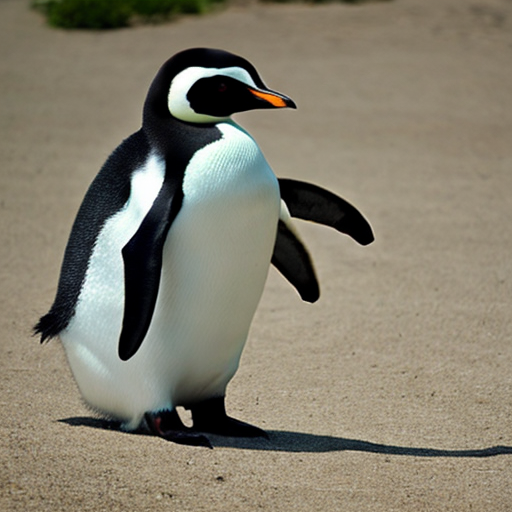}
    \end{tabular}
    \caption{We present images obtained with the same seed and same prompt for SD 1.4 (left) and SD 2 (right). Note how similar the compositions are. The prompt and seed are respectively "\textit{a photo of a blue circle}" seed 17, "\textit{a photo of a cat}" seed 27, "\textit{a photo of a pink elephant}" seed 18, and "\textit{a photo of a penguin}" seed 45.}
    \label{fig:sd_2_1_4}
\end{figure*}

As the two models are trained on the same data, we hypothesize that they are trained on the same $z_t$ \ie they are trained to predict the same $\epsilon$. However $\epsilon$ is random, we suspect that all the $z_t$ are precomputed and the associate $\epsilon$ saved, and they used the same for the training of both models. Because of this particularity during the training, the models try to retrieve the $z_0$ from multiple possible $t$. The models learn a similar path of reverse diffusion. At inference, when we draw a similar noise $\chi_T$, the reverse process will be similar and conduct to near $\chi_t$ and finally to a near $\chi_0$. We show in Fig.~\ref{fig:sd_2_1_4} that with the same prompt and the same starting noise, we exhibit strong composition similarity that explains the parallel performance of the seed.

\newpage
\section{Human Evaluation and Comparison to Other Automatic Metrics} \label{appendix:compare_human_other}

We conducted a human evaluation to determine how much TIAM is aligned to it, and compared it to two other automatic metrics based on the CLIP\cite{radford2021learning} score and BLIP\cite{li2022blip} score.

\begin{figure*}
    \centering
    \includegraphics[width=0.8\linewidth]{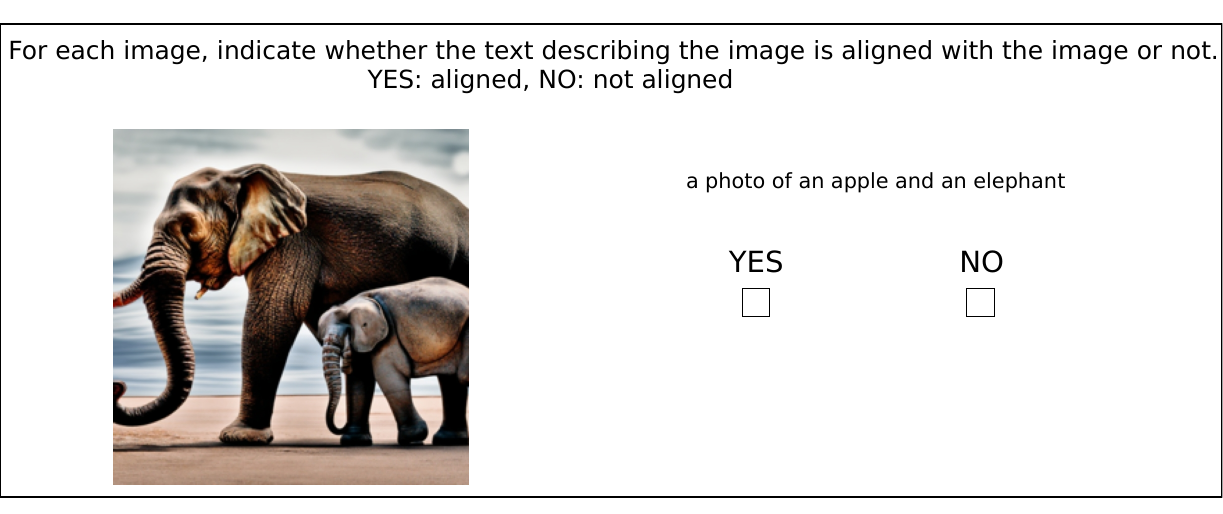}
    \caption{Extract from the study for the human evaluation.}
    \label{fig:human_study}
\end{figure*}

We randomly sample 32 prompts, comprising 16 with one object and one attribute/color (referred to as $\mathcal{C}$), and 16 with two objects (referred to as $\mathcal{O}$). Subsequently, we randomly select one image per prompt and record the corresponding scores provided by our metrics for each image. We use images generated by the IF model. We then solicit human evaluators to discern whether they perceive alignment between the given prompt (utilized for generation) and the associated image. An illustrative example is given in Fig.~\ref{fig:human_study}. For a fair comparison with CLIP and BLIP, which solely yield a similarity score between images and captions, we refrained from rephrasing the prompts into specific questions like ''Is the first object present?’’ or ''Is the second object present?’’, that would have given an advantage to TIAM.

We conducted our study on 57 humans aged from 7 to 79. Only 6 of them could be considered text-to-image (T2I) experts (the author who made the study did not participate in the assessment), while other human subjects never manipulated T2I models, or even didn't know it could exist. In any case, the agreement of TIAM with the experts' assessment was not significantly different than that with non-experts. Nor did we find any significant difference in terms of gender or age. For non-English speakers, the prompts were translated into their native language (in particular for subjects less than 15 years old).

For automatic methods, we used the OpenAI ViT-B/32 CLIP model \footnote{\href{https://huggingface.co/openai/clip-vit-base-patch32}{https://huggingface.co/openai/clip-vit-base-patch32}} and the Salesforce BLIP model \footnote{\href{https://huggingface.co/Salesforce/blip-itm-base-coco}{https://huggingface.co/Salesforce/blip-itm-base-coco}}. The BLIP and CLIP score is a similarity score between the embedding of the image and the caption used to generate the image.

\begin{figure}[]
    \centering
    \includegraphics[width=0.7\linewidth]{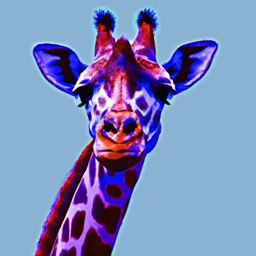}
    \caption{``A photo of a blue giraffe'' generated with IF. The human annotators had quite low agreement on the alignment of this image with the prompt.}
    \label{fig:a_blue_giraffe}
\end{figure}

The agreement between the human annotator was assessed in terms of Fleiss' kappa~\cite{fleiss1971agreement}. According to Landis and Koch ~\cite{landis1977agreement}, a kappa of $[0.21-0.40]$ is \textit{fair}, that in $[0.41-0.60]$ is \textit{moderate}, that in $[0.61-0.80]$ is \textit{substantial} and the agreement is \textit{almost perfect} when the Fleiss' kappa is in $[0.81-1.00]$. With a global value of  $\kappa = 0.73$ the agreement of the human annotators of our study is thus \textit{substantial}. If one distinguishes the two subsets, the agreement on $\mathcal{C}$ is in the upper range of the \textit{moderate} agreement ($0.59$) while that for $\mathcal{O}$ is \textit{almost perfect} ($0.85$). It nevertheless shows that, even for humans, characterizing the colors of an object may be an ambiguous task. We illustrate for instance the example that led to the most disagreement between annotators in Fig.~\ref{fig:a_blue_giraffe}.

We compute the Pearson correlation between human decisions and TIAM, the CLIP score, and the BLIP score. The results are reported in Tab.~\ref{tab:correlation_pearson}. TIAM exhibits a significantly stronger correlation with human judgments compared with other metrics. We have a similar conclusion if the alignment is estimated with the Spearman'rank correlation (Tab.~\ref{tab:correlation_spearman}).

\begin{table}[]
    {\small{\begin{center}
                \begin{tabular}{llll}
                    \toprule
                    Score & $\mathcal{C}+\mathcal{O}$            & $\mathcal{C}$                        & $\mathcal{O}$                         \\
                    \midrule
                    CLIP  & $0.47_{p=6\times 10^{-3}}$           & $0.22_{p=4\times 10^{-1}}$           & $0.62_{p=1 \times 10^{-2}}$           \\
                    BLIP  & $0.67_{p=3 \times 10^{-5}}$          & $0.48_{p=6 \times 10^{-2}}$          & $0.77_{p=5 \times 10^{-4}}$           \\
                    TIAM  & $\mathbf{0.82}_{p=7 \times 10^{-9}}$ & $\mathbf{0.70}_{p=2 \times 10^{-3}}$ & $\mathbf{0.98}_{p=2 \times 10^{-11}}$ \\
                    \bottomrule
                \end{tabular}
            \end{center}}}
    \caption{Pearson correlation between human decisions and TIAM/BLIP/CLIP. $\mathcal{C}+\mathcal{O}$ stands for the correlation without distinction of the series. $p$ is the p-value for the null hypothesis $H_0$: \textit{the distributions underlying the samples are uncorrelated}, the alternative hypothesis is \textit{the correlation is non zero}.}
    \label{tab:correlation_pearson}
\end{table}

\begin{table}[]
    {\small{\begin{center}
                \begin{tabular}{llll}
                    \toprule
                    Score & $\mathcal{C}+\mathcal{O}$           & $\mathcal{C}$                       & $\mathcal{O}$                       \\
                    \midrule
                    CLIP  & $0.50_{p=2\times 10^{-3}}$          & $0.39_{p=7\times 10^{-2}}$          & $0.53_{p=2\times 10^{-02}}$         \\
                    BLIP  & $0.38_{p=2\times 10^{-2}}$          & $0.06_{p=4\times 10^{-1}}$          & $0.64_{p=4\times 10^{-3}}$          \\
                    TIAM  & $\mathbf{0.82}_{p=1\times 10^{-4}}$ & $\mathbf{0.77}_{p=1\times 10^{-3}}$ & $\mathbf{0.87}_{p=2\times 10^{-4}}$ \\
                    \bottomrule
                \end{tabular} 
            \end{center}}}
    \caption{Spearman correlation between human decisions and TIAM/BLIP/CLIP. $\mathcal{C}+\mathcal{O}$ stands for the correlation without distinction of the series. $p$ is the p-value for the permutation test.}
    \label{tab:correlation_spearman}
\end{table}

Lastly, we would like to emphasize that TIAM captures the model's success rate more comprehensibly in contrast to other automatic similarity scores. While CLIP and BLIP can serve to compare two generative models (\eg evaluate models based on their CLIP score) their inherent meaning is limited. We note also that CLIP has a poor compositional understanding,  limiting a precise evaluation of text-image alignment \cite{yuksekgonul2023when}. In addition, TIAM enables the analysis of specific modalities, yielding insightful outcomes such as the success rate per seed and the proportion of apparition of an object according to its position in the prompt.

\section{Scalability of TIAM}

In this section, we initially address strategies to manage the potential growing complexity inherent in the template approach. Subsequently, we explore methods to go beyond the limited set of COCO labels.

\subsection{Scalability\label{appendix:scalabilityTiam}}

The template approach can become cumbersome when dealing with multiple modalities (\eg exploring a prompt template with 5 objects using 30 different objects ($\mathcal{0}| = 30$) leads to 17 100 720 prompts).

To alleviate this complexity, we can adopt a sampling approach, consisting of randomly drawing a defined number of prompts to estimate the results (\eg TIAM score, the proportion of occurrence of the object according to its position in the template, \dots). We conducted such a study on several experiments reported in the paper :
(A) the prompts with 2 objects created with the combination of 24 COCO labels (Section~\ref{sec:preliminary}), (B) the prompts with 2 objects created with the combination of 28 COCO labels (Section~\ref{sec:catastrophic_neglect}, part on the semantic link), and (C) the prompts with 2 objects and associated attribute (Section~\ref{sec:4_4_attributeBinding}).

\begin{figure*}
    \centering
    \includegraphics[width=0.8\linewidth]{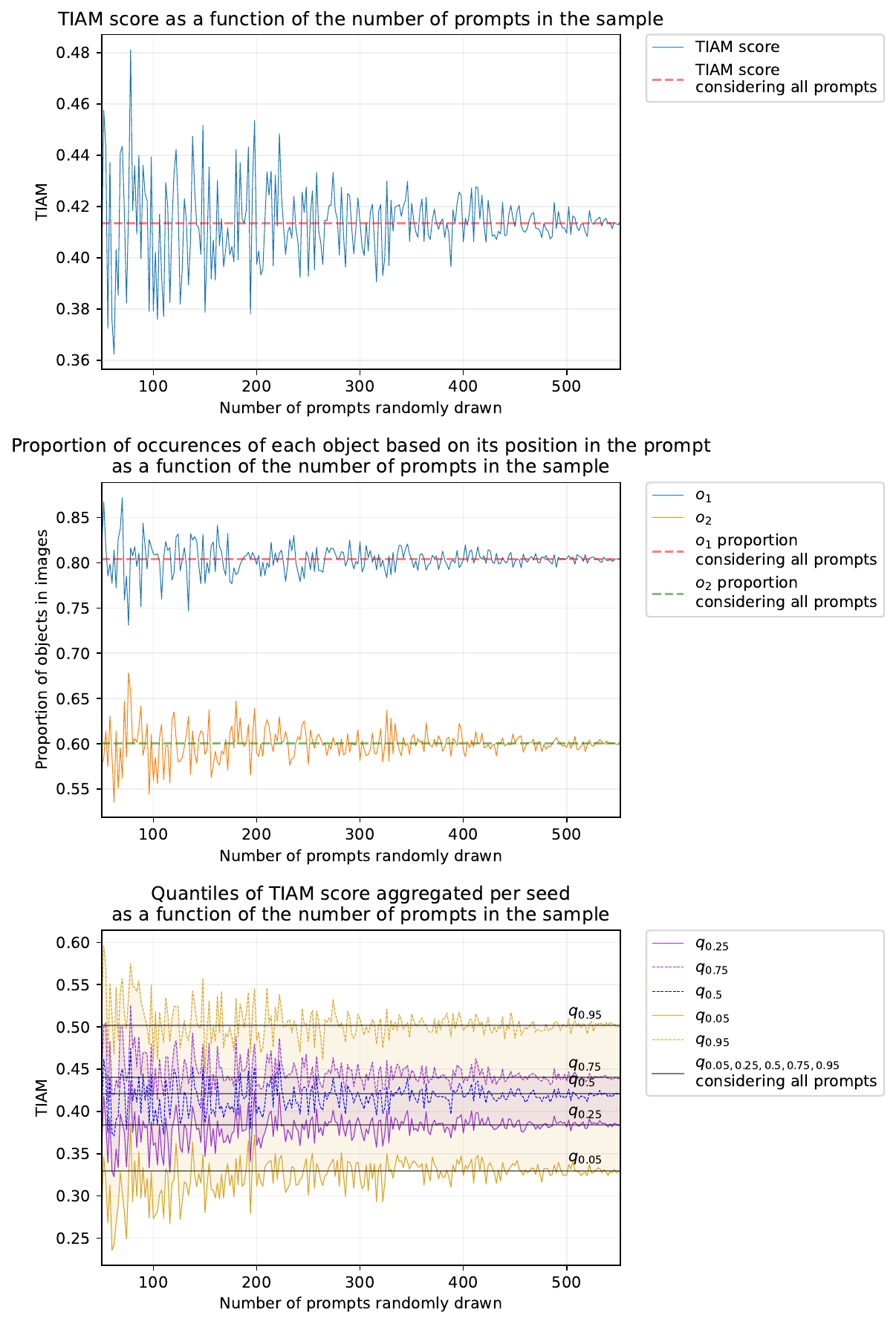}
    \caption{Evolution of, respectively, the TIAM score, the proportion of occurrences of each object based on its position in the prompt, and the quantiles of the TIAM score aggregated per seed as a function of the number of prompts randomly drawn to compute the results, for SD 1.4, using the prompts with 2 objects created with the combination of 24 COCO labels (Section~\ref{sec:preliminary}).}
    \label{fig:scalability_2o_1_4}
\end{figure*}

\begin{figure*}
    \centering
    \includegraphics[width=0.8\linewidth]{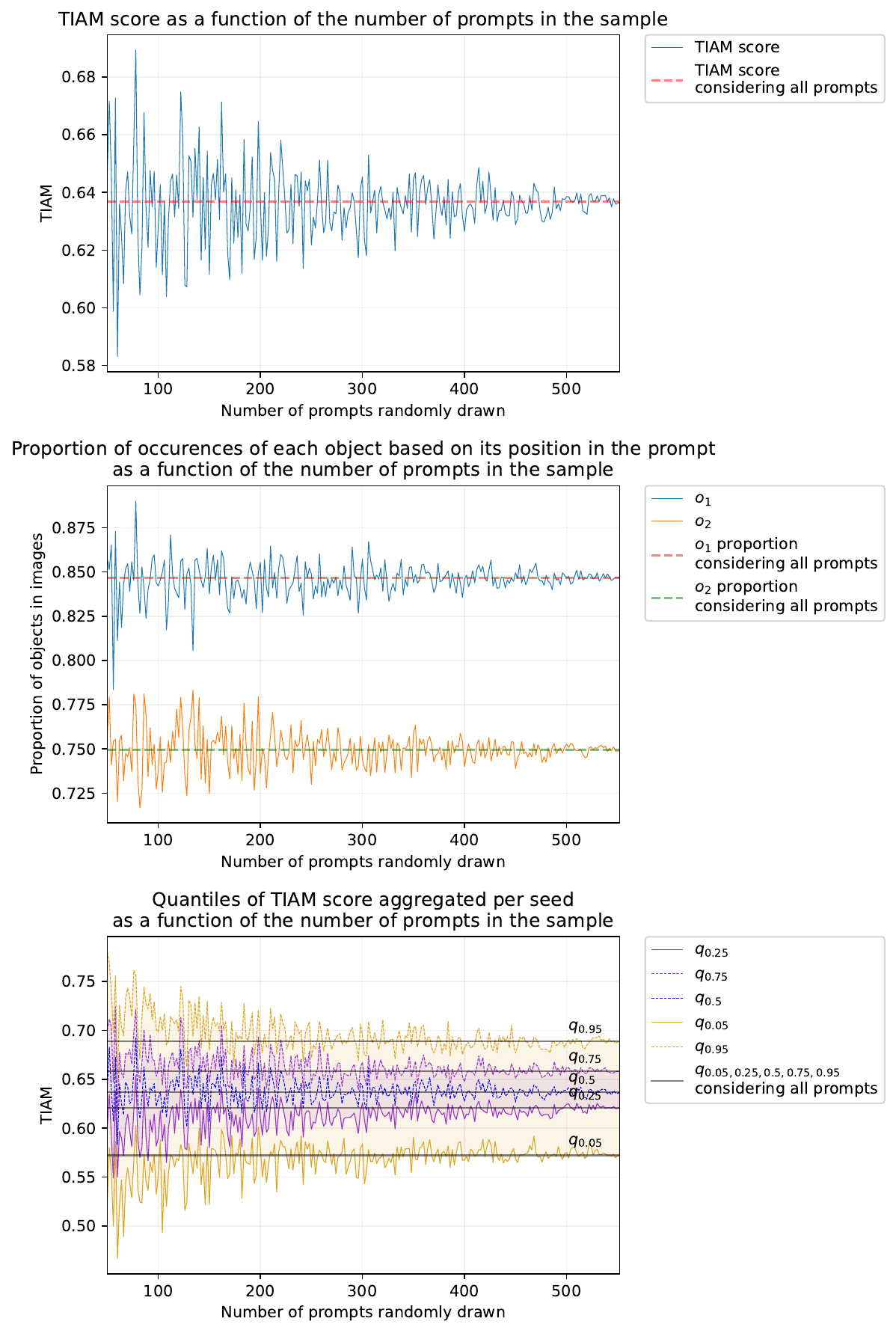}
    \caption{Evolution of, respectively, the TIAM score, the proportion of occurrences of each object based on its position in the prompt, and the quantiles of the TIAM score aggregated per seed as a function of the number of prompts randomly drawn to compute the results, for SD 1.4 A\&E, using the prompts with 2 objects created with the combination of 24 COCO labels (Section~\ref{sec:preliminary}).}
    \label{fig:scalability_2o_1_4_ae}
\end{figure*}

\begin{figure*}
    \centering
    \includegraphics[width=0.8\linewidth]{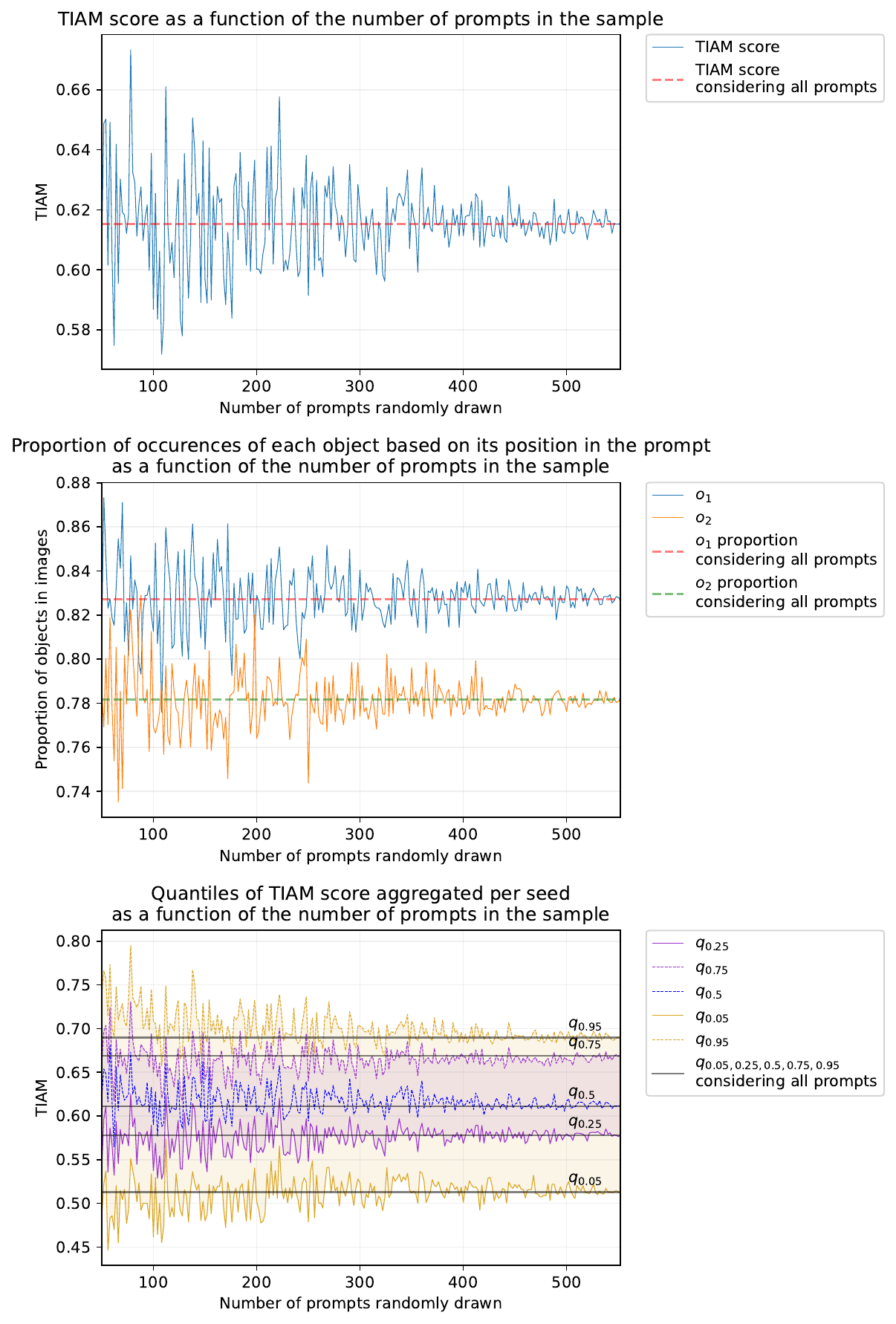}
    \caption{Evolution of, respectively, the TIAM score, the proportion of occurrences of each object based on its position in the prompt, and the quantiles of the TIAM score aggregated per seed as a function of the number of prompts randomly drawn to compute the results, for SD 2, using the prompts with 2 objects created with the combination of 24 COCO labels (Section~\ref{sec:preliminary}).}
    \label{fig:scalability_2o_v2}
\end{figure*}

\begin{figure*}
    \centering
    \includegraphics[width=0.8\linewidth]{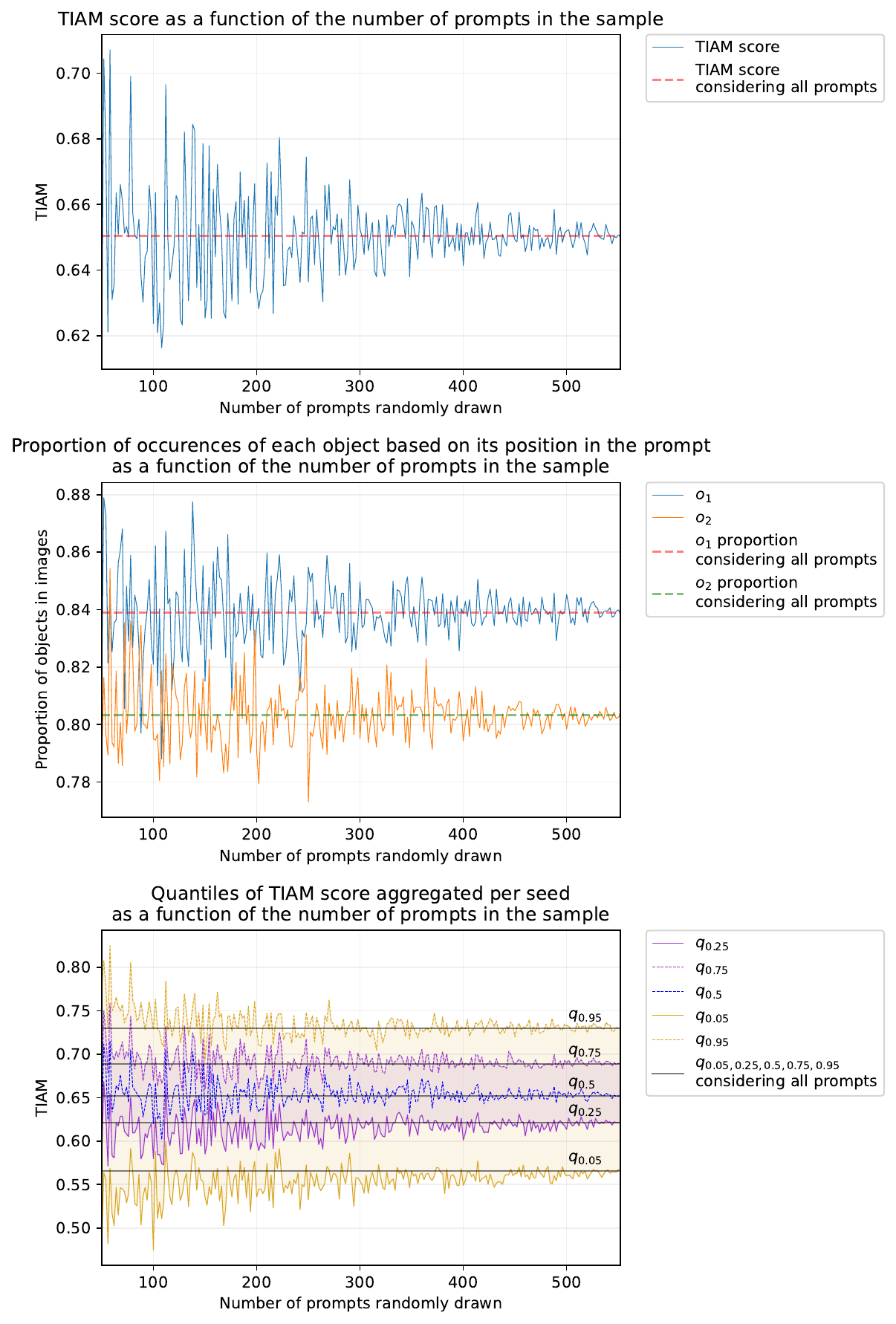}
    \caption{Evolution of, respectively, the TIAM score, the proportion of occurrences of each object based on its position in the prompt, and the quantiles of the TIAM score aggregated per seed as a function of the number of prompts randomly drawn to compute the results, for SD 2 A\&E, using the prompts with 2 objects created with the combination of 24 COCO labels (Section~\ref{sec:preliminary}).}
    \label{fig:scalability_2o_v2_ae}
\end{figure*}

\begin{figure*}
    \centering
    \includegraphics[width=0.8\linewidth]{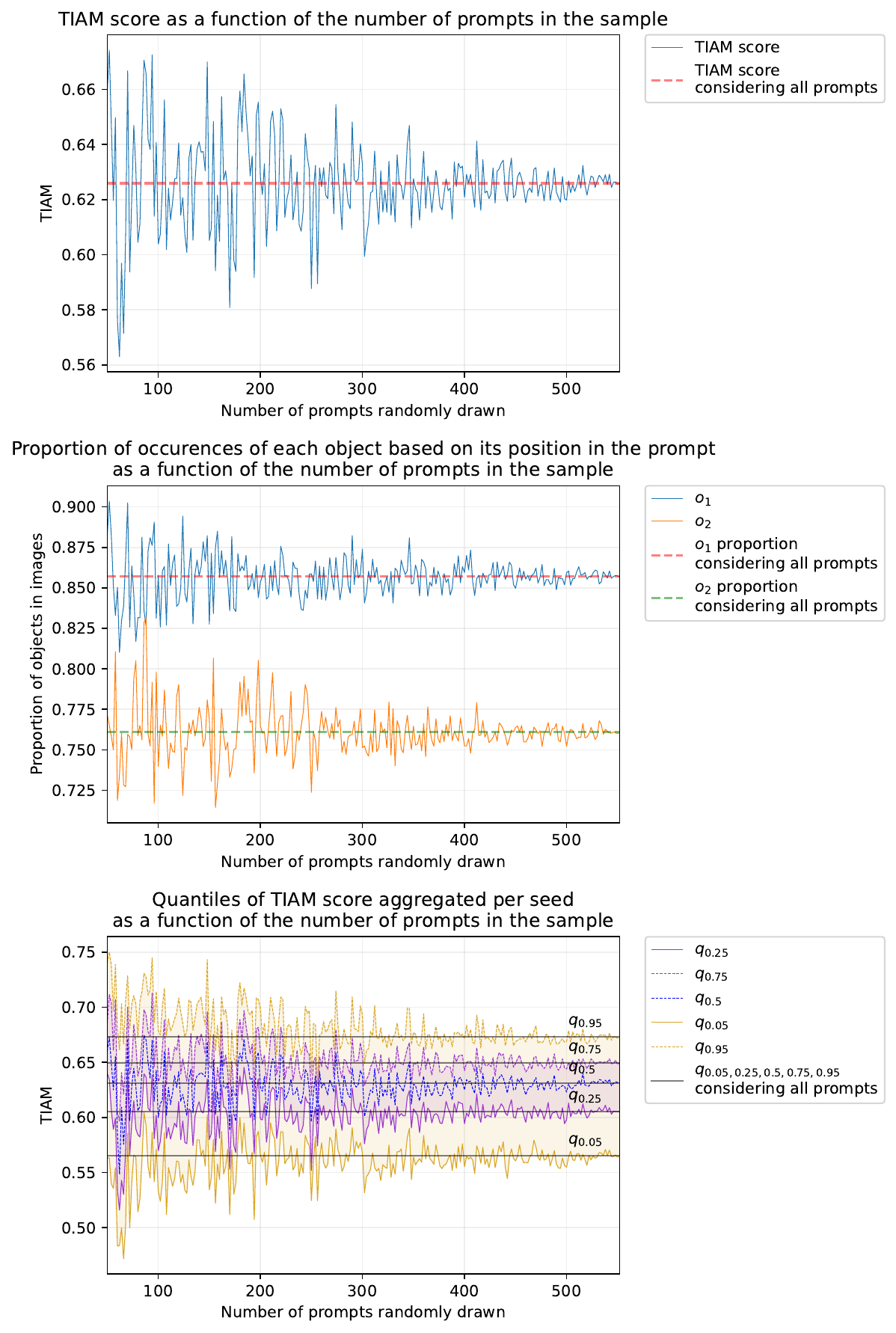}
    \caption{Evolution of, respectively, the TIAM score, the proportion of occurrences of each object based on its position in the prompt, and the quantiles of the TIAM score aggregated per seed as a function of the number of prompts randomly drawn to compute the results, for IF, using the prompts with 2 objects created with the combination of 24 COCO labels (Section~\ref{sec:preliminary}).}
    \label{fig:scalability_2o_if}
\end{figure*}

\begin{figure*}
    \centering
    \includegraphics[width=0.8\linewidth]{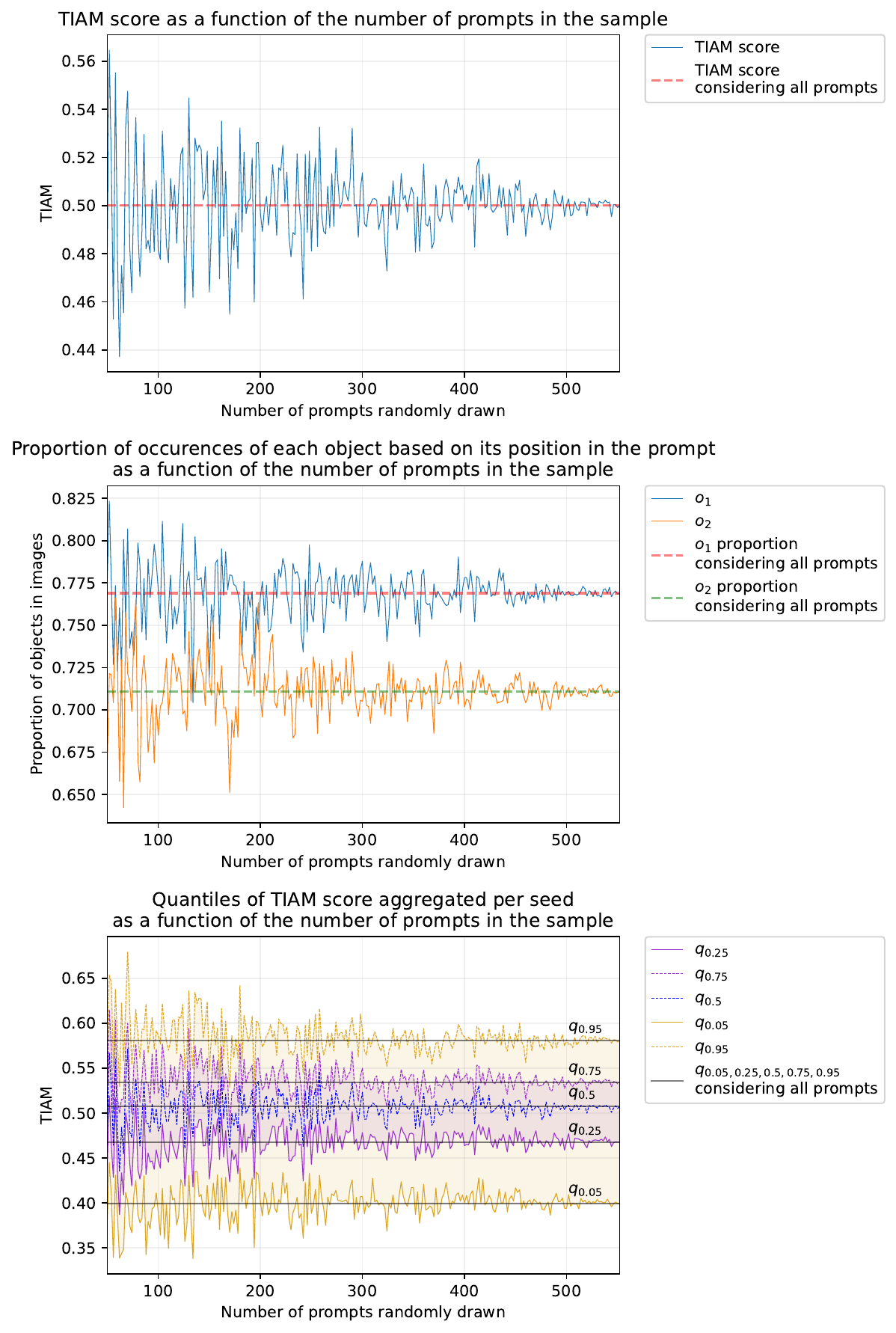}
    \caption{Evolution of, respectively, the TIAM score, the proportion of occurrences of each object based on its position in the prompt, and the quantiles of the TIAM score aggregated per seed as a function of the number of prompts randomly drawn to compute the results, for the unCLIP, using the prompts with 2 objects created with the combination of 24 COCO labels (Section~\ref{sec:preliminary}).}
    \label{fig:scalability_2o_unclip}
\end{figure*}

\begin{figure*}
    \centering
    \includegraphics[width=0.8\linewidth]{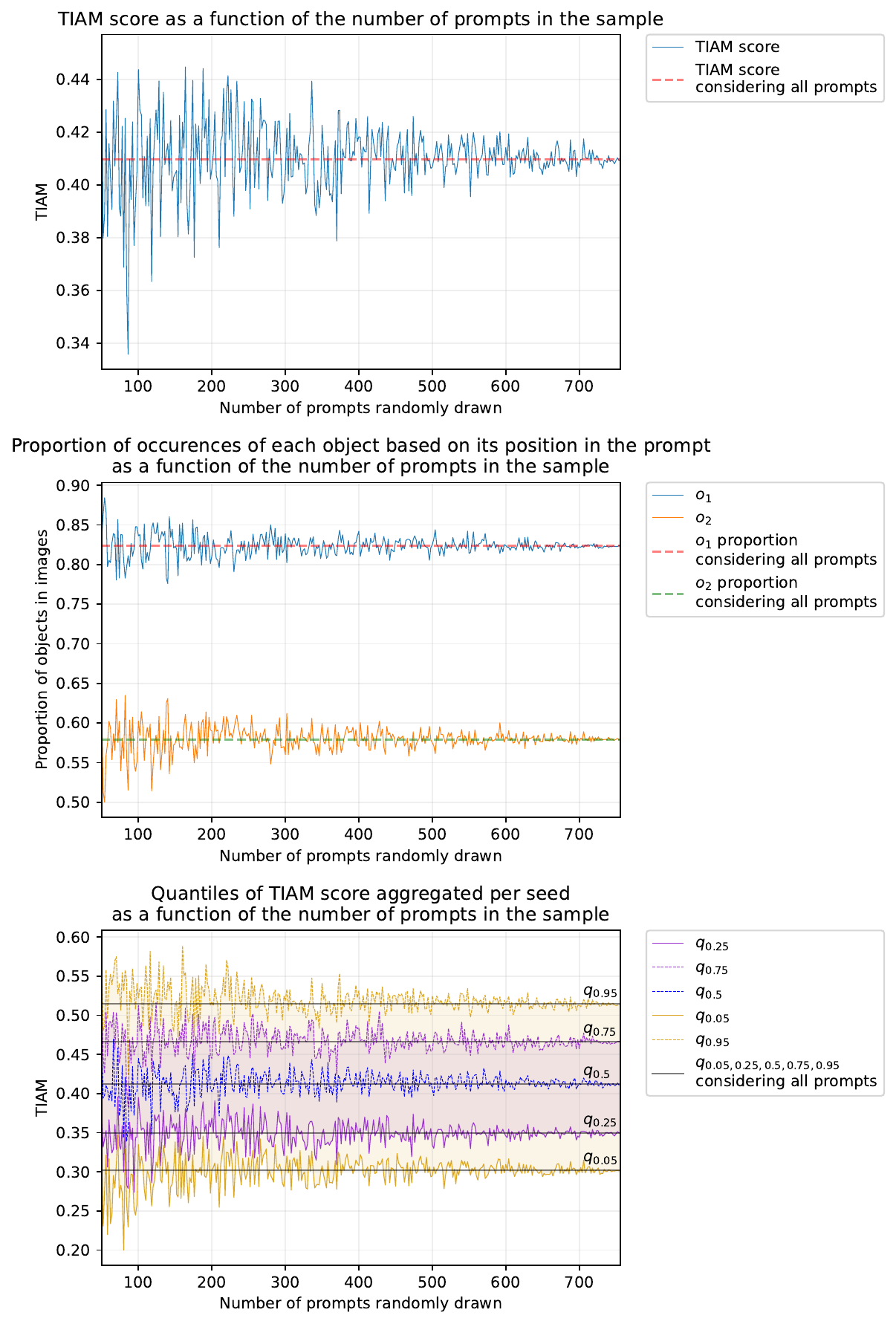}
    \caption{Evolution of, respectively, the TIAM score, the proportion of occurrences of each object based on its position in the prompt, and the quantiles of the TIAM score aggregated per seed as a function of the number of prompts randomly drawn to compute the results, for SD 1.4, using the prompts with 2 objects created with the combination of 28 COCO labels (Section~\ref{sec:catastrophic_neglect}, part on the semantic link).}
    \label{fig:scalability_28_2o_1_4}
\end{figure*}

\begin{figure*}
    \centering
    \includegraphics[width=0.8\linewidth]{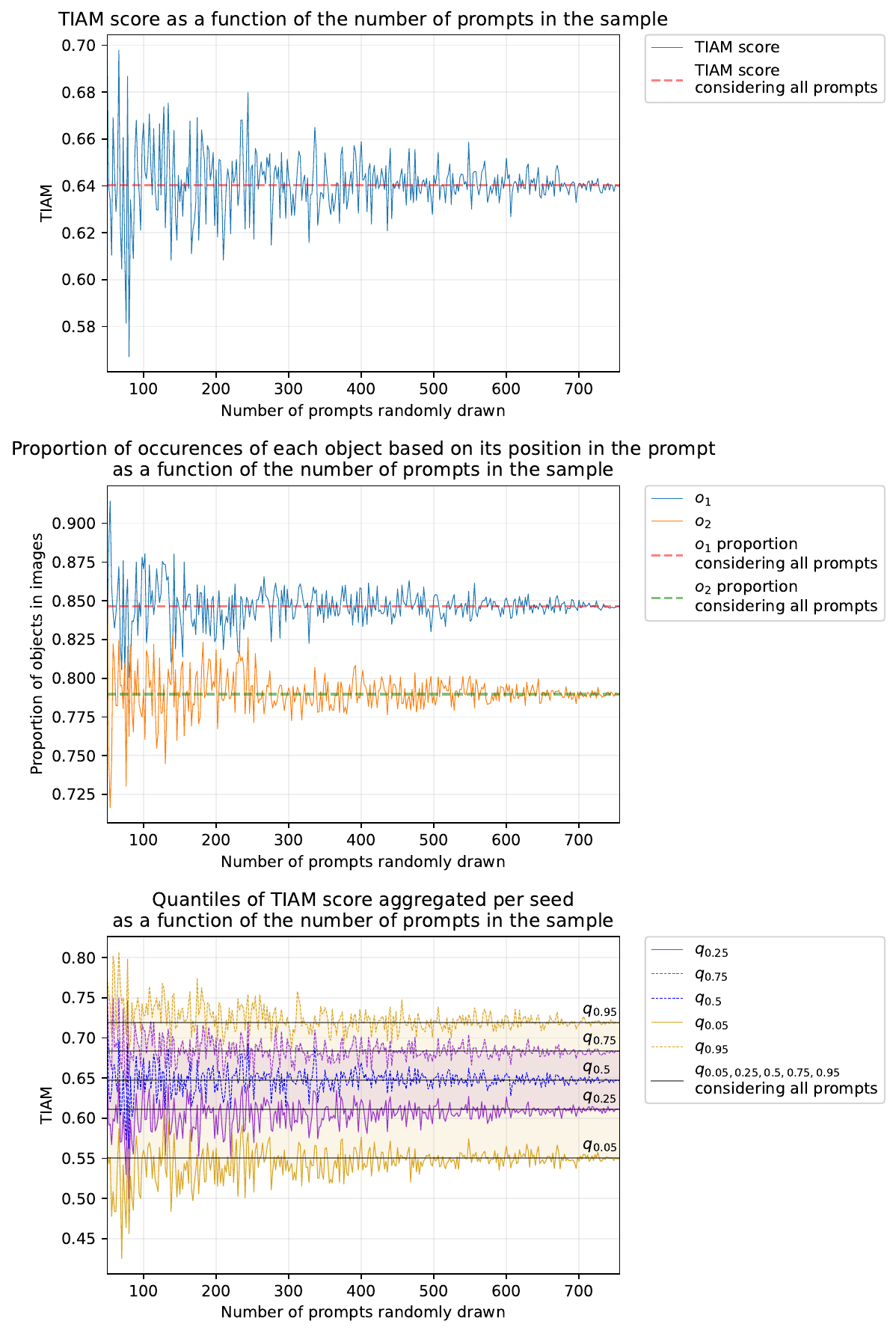}
    \caption{Evolution of, respectively, the TIAM score, the proportion of occurrences of each object based on its position in the prompt, and the quantiles of the TIAM score aggregated per seed as a function of the number of prompts randomly drawn to compute the results, for SD 2, using the prompts with 2 objects created with the combination of 28 COCO labels (Section~\ref{sec:catastrophic_neglect}, part on the semantic link).}
    \label{fig:scalability_28_2o_v2}
\end{figure*}

\begin{figure*}
    \centering
    \includegraphics[width=0.8\linewidth]{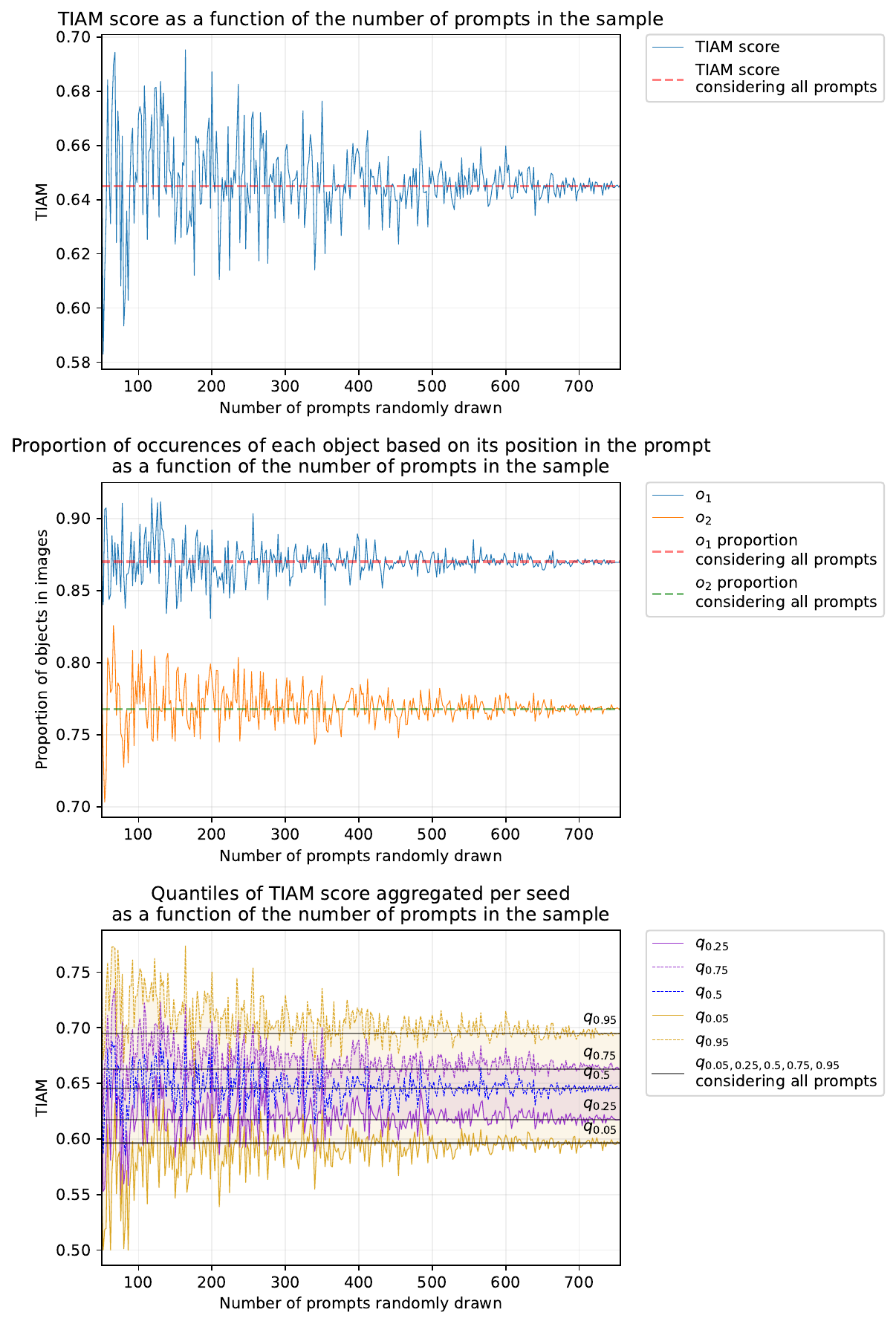}
    \caption{Evolution of, respectively, the TIAM score, the proportion of occurrences of each object based on its position in the prompt, and the quantiles of the TIAM score aggregated per seed as a function of the number of prompts randomly drawn to compute the results, for IF, using the prompts with 2 objects created with the combination of 28 COCO labels (Section~\ref{sec:catastrophic_neglect}, part on the semantic link).}
    \label{fig:scalability_28_2o_if}
\end{figure*}

\begin{figure*}
    \centering
    \includegraphics[width=0.8\linewidth]{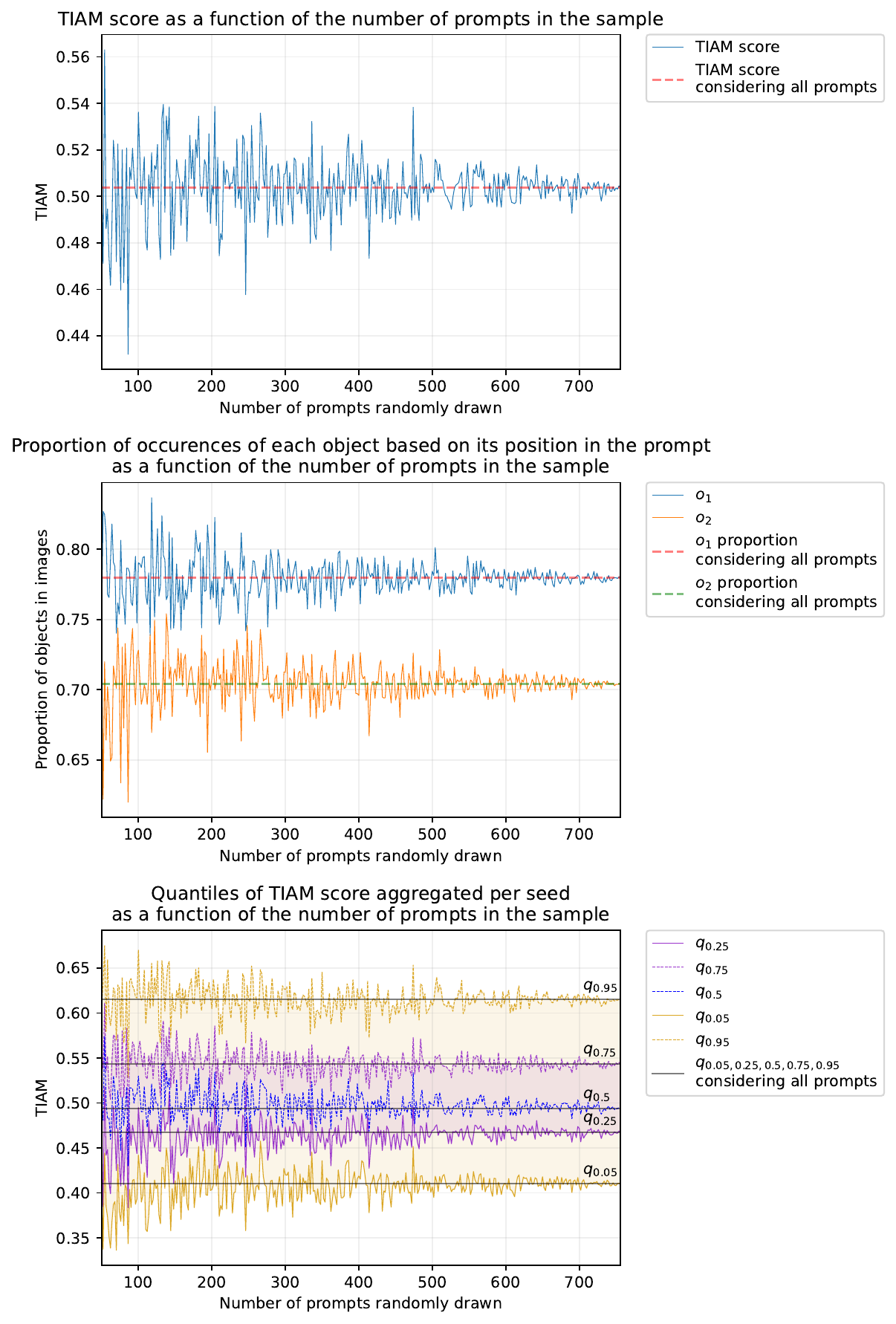}
    \caption{Evolution of, respectively, the TIAM score, the proportion of occurrences of each object based on its position in the prompt, and the quantiles of the TIAM score aggregated per seed as a function of the number of prompts randomly drawn to compute the results, for unCLIP, using the prompts with 2 objects created with the combination of 28 COCO labels (Section~\ref{sec:catastrophic_neglect}, part on the semantic link).}
    \label{fig:scalability_28_2o_unclip}
\end{figure*}

\begin{figure*}
    \centering
    \includegraphics[width=0.8\linewidth]{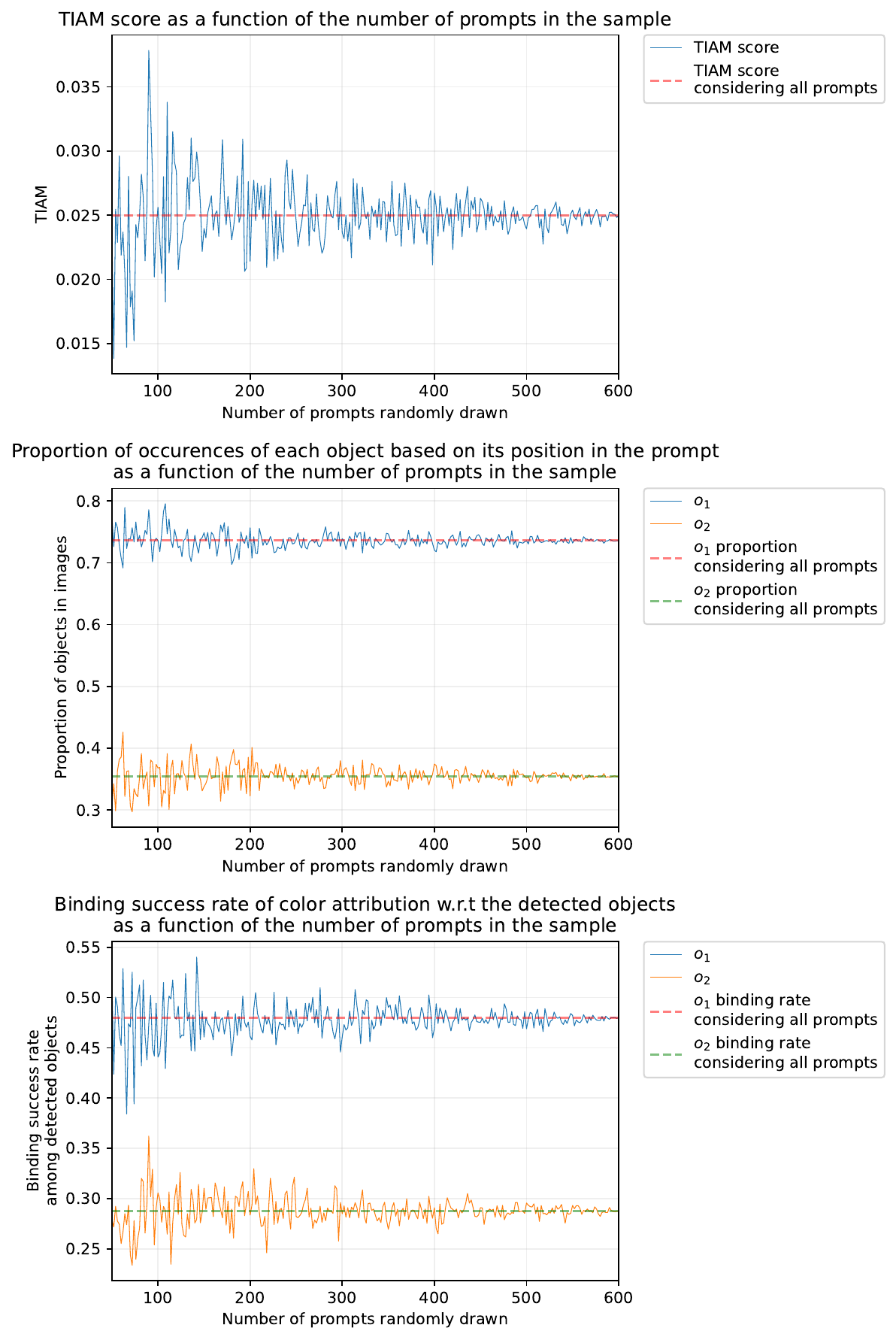}
    \caption{Evolution of, respectively, the TIAM score, the proportion of occurrences of each object based on its position in the prompt, and the success rate of color attribution w.r.t the detected objects as a function of the number of prompts randomly drawn to compute the results, for SD 1.4, using the prompts with 2 objects and associated attribute (Section~\ref{sec:4_4_attributeBinding}).}
    \label{fig:scalability_colors_1_4}
\end{figure*}

\begin{figure*}
    \centering
    \includegraphics[width=0.8\linewidth]{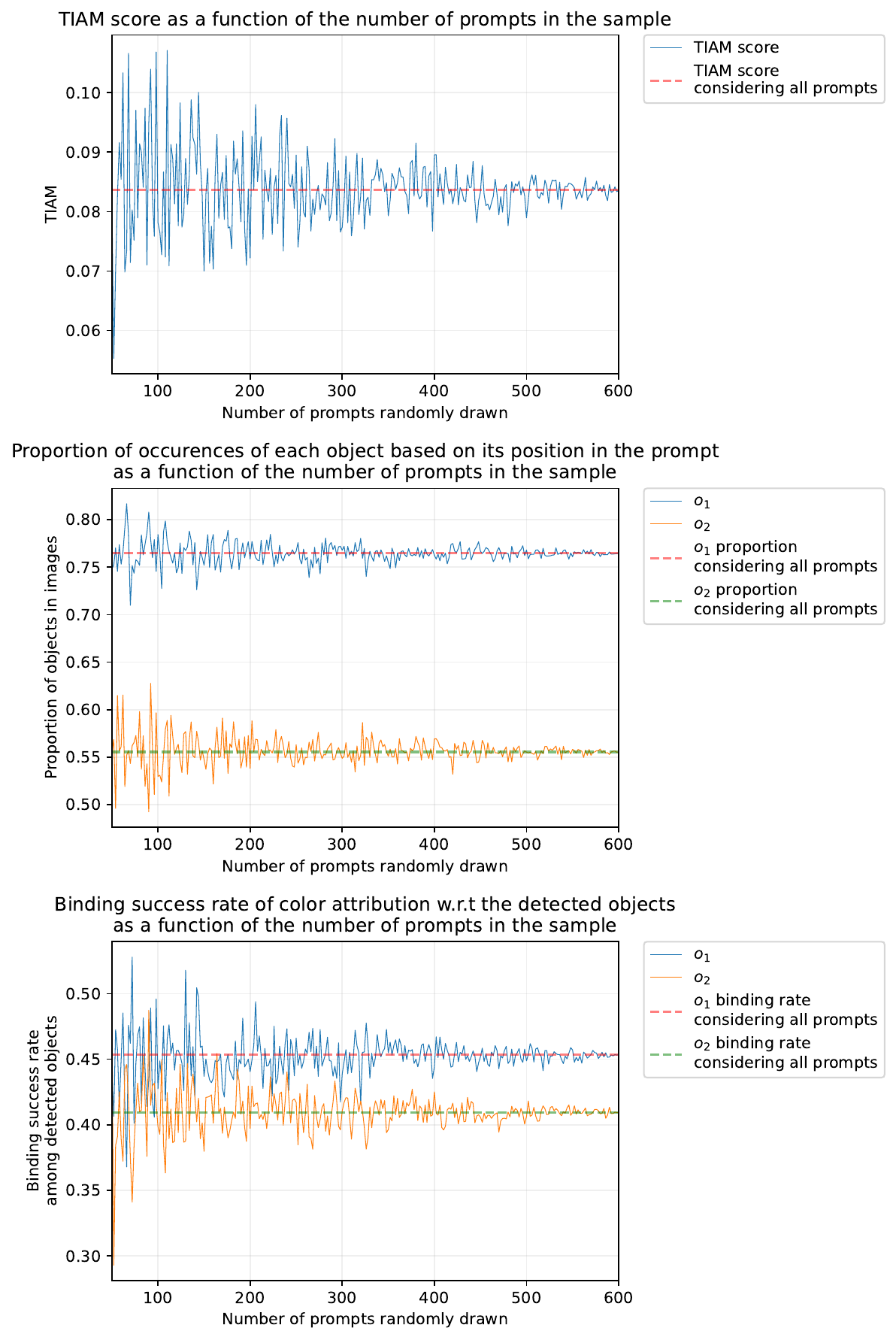}
    \caption{Evolution of, respectively, the TIAM score, the proportion of occurrences of each object based on its position in the prompt, and the success rate of color attribution w.r.t the detected objects as a function of the number of prompts randomly drawn to compute the results, for SD 1.4 A\&E, using the prompts with 2 objects and associated attribute (Section~\ref{sec:4_4_attributeBinding}).}
    \label{fig:scalability_colors_1_4_ae}
\end{figure*}

\begin{figure*}
    \centering
    \includegraphics[width=0.8\linewidth]{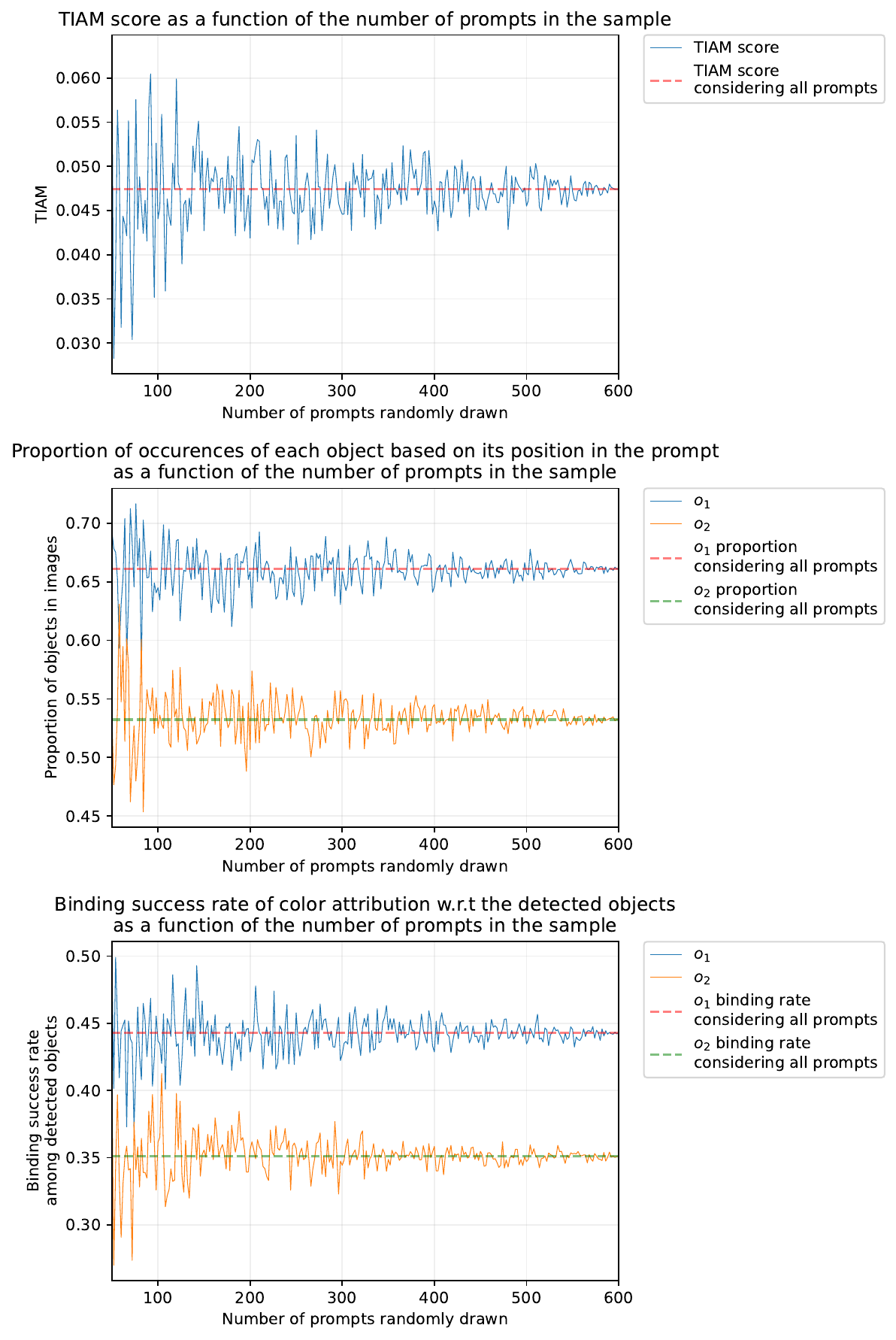}
    \caption{Evolution of, respectively, the TIAM score, the proportion of occurrences of each object based on its position in the prompt, and the success rate of color attribution w.r.t the detected objects as a function of the number of prompts randomly drawn to compute the results, for SD 2, using the prompts with 2 objects and associated attribute (Section~\ref{sec:4_4_attributeBinding}).}
    \label{fig:scalability_colors_2}
\end{figure*}

\begin{figure*}
    \centering
    \includegraphics[width=0.8\linewidth]{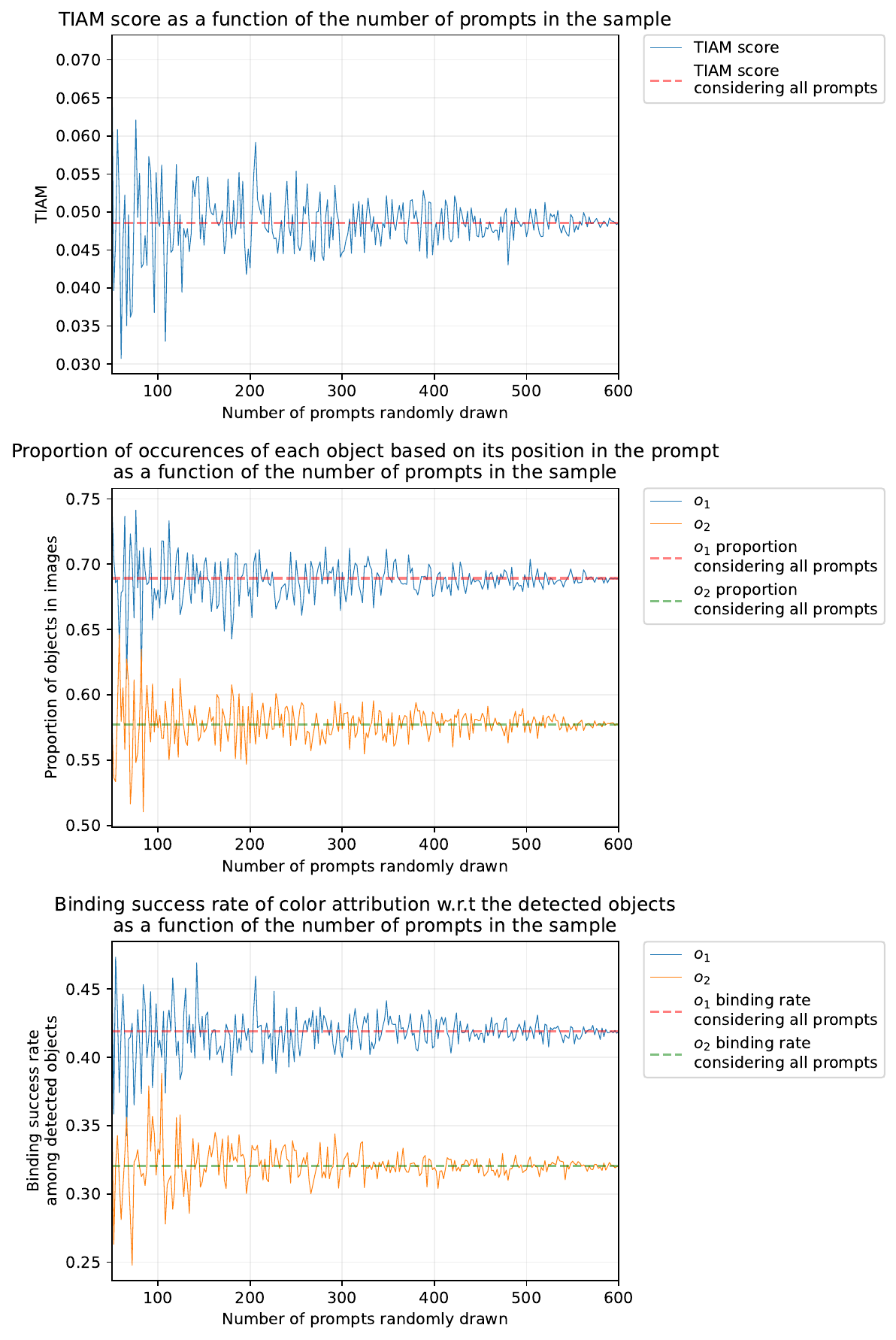}
    \caption{Evolution of, respectively, the TIAM score, the proportion of occurrences of each object based on its position in the prompt and the success rate of color attribution w.r.t the detected objects as a function of the number of prompts randomly drawn to compute the results, for SD 2 A\&E, using the prompts with 2 objects and associated attribute (Section~\ref{sec:4_4_attributeBinding}).}
    \label{fig:scalability_colors_2_ae}
\end{figure*}

\begin{figure*}
    \centering
    \includegraphics[width=0.8\linewidth]{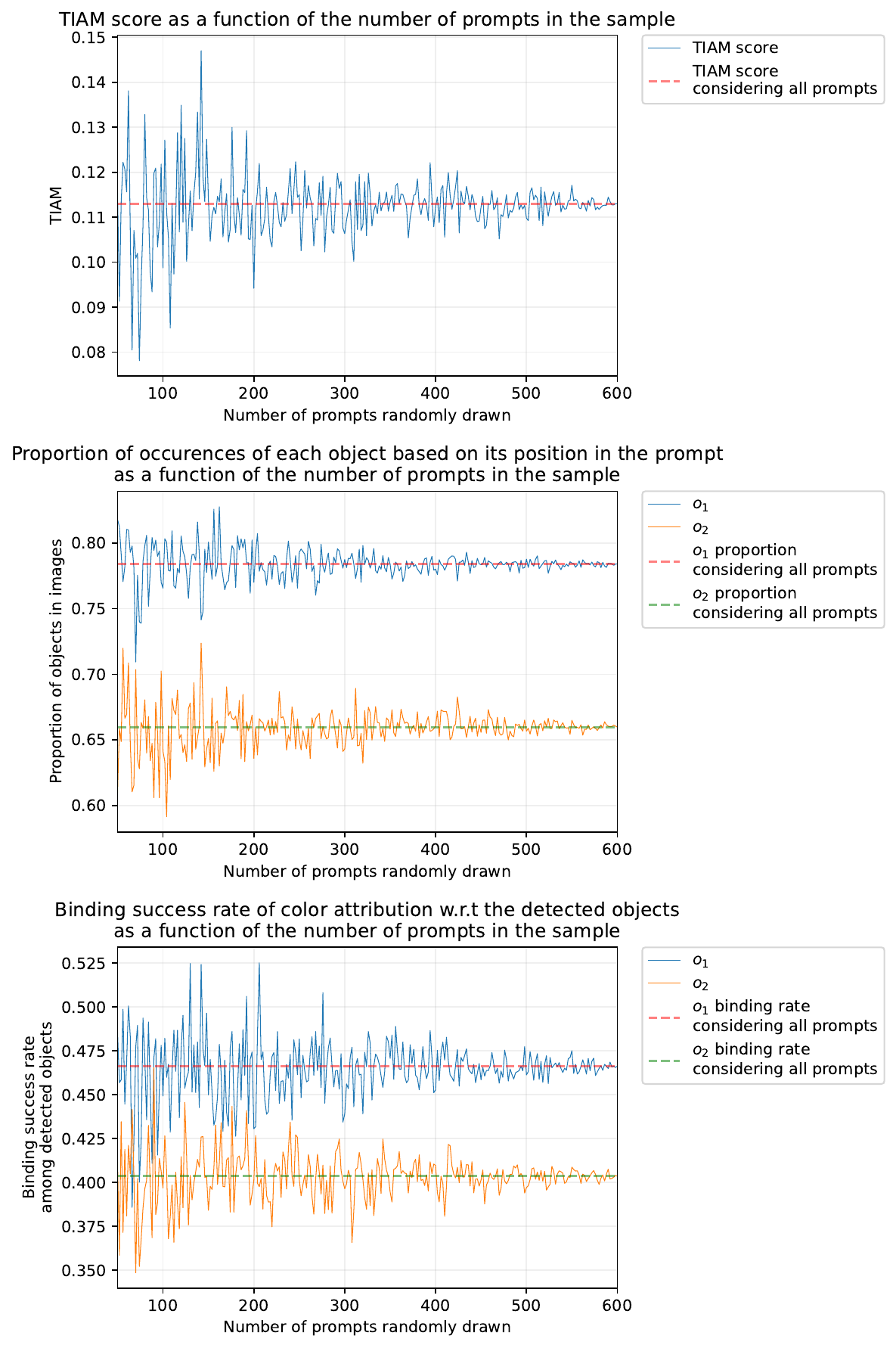}
    \caption{Evolution of, respectively, the TIAM score, the proportion of occurrences of each object based on its position in the prompt, and the success rate of color attribution w.r.t the detected objects as a function of the number of prompts randomly drawn to compute the results, for IF, using the prompts with 2 objects and associated attribute (Section~\ref{sec:4_4_attributeBinding}).}
    \label{fig:scalability_colors_if}
\end{figure*}

\begin{figure*}
    \centering
    \includegraphics[width=0.8\linewidth]{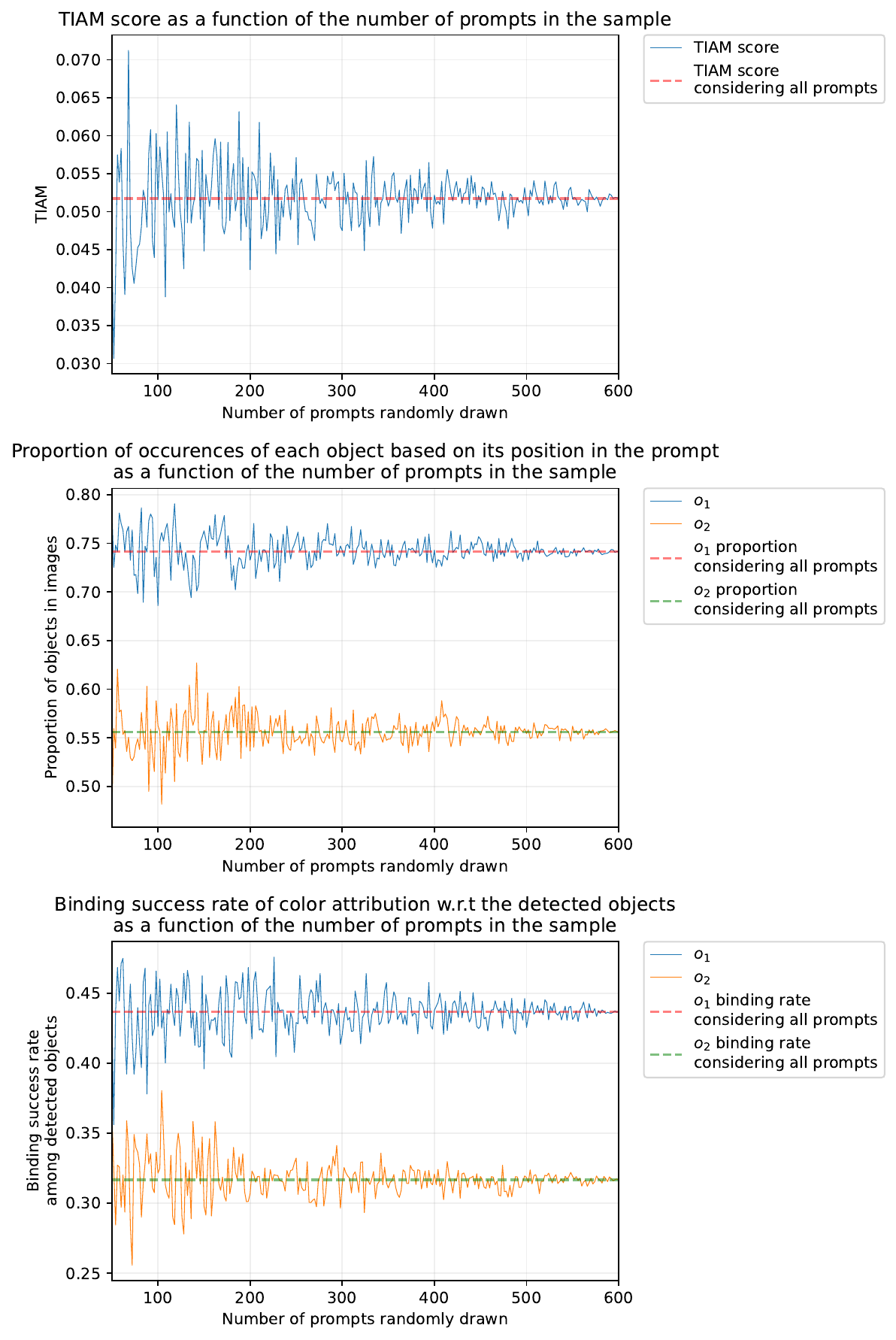}
    \caption{Evolution of, respectively, the TIAM score, the proportion of occurrences of each object based on its position in the prompt, and the success rate of color attribution w.r.t the detected objects as a function of the number of prompts randomly drawn to compute the results, for unCLIP, using the prompts with 2 objects and associated attribute (Section~\ref{sec:4_4_attributeBinding}).}
    \label{fig:scalability_colors_unclip}
\end{figure*}

Using all generated prompts, at each step, we draw (without replacement) $n$ prompts and we compute the results from these samples. We start with 50 prompts and increase $n$ up to the maximum number (all possible prompts) by a step of 2 (thus using $52, 54,56...$ samples). We made this for each model used for the respective experiments. For (A) and (B) we report the TIAM score, the proportion of occurrences of each object based on its position in the prompt, and the quantiles of the TIAM score aggregated per seed, for (C) we report the TIAM score,  the proportion of occurrences of each object based on its position in the prompt and the success rate of color attribution w.r.t the detected objects. Hence we report below the results for:
\begin{itemize}
    \item SD 1-4 (A) Fig.~\ref{fig:scalability_2o_1_4} (B) Fig.~\ref{fig:scalability_28_2o_1_4} (C) Fig.~\ref{fig:scalability_colors_1_4},
    \item SD 1.4 A\&E (A) Fig.~\ref{fig:scalability_2o_1_4_ae} (C) Fig.~\ref{fig:scalability_colors_1_4_ae},
    \item SD 2 (A) Fig.~\ref{fig:scalability_2o_v2} (B) Fig.~\ref{fig:scalability_28_2o_v2} (C) Fig.~\ref{fig:scalability_colors_2},
    \item SD 2 A\&E (A) Fig.~\ref{fig:scalability_2o_v2_ae} (C) Fig.~\ref{fig:scalability_colors_2_ae},
    \item IF (A) Fig.~\ref{fig:scalability_2o_if} (B) Fig.~\ref{fig:scalability_28_2o_if} (C) Fig.~\ref{fig:scalability_colors_if},
    \item unCLIP (A) Fig.~\ref{fig:scalability_2o_unclip} (B) Fig.~\ref{fig:scalability_28_2o_unclip} (C) Fig.~\ref{fig:scalability_colors_unclip}.
\end{itemize}
Across all our findings, a marked trend emerges from around 300 prompts, indicating the viability of employing a sampling method to approximate the results presented in the main study and alleviate the complexity of the template-based approach of TIAM in practice.

In addition, modal-specific studies can be conducted. For instance, to study certain modalities, we can imagine isolating each modality for individual scrutiny such as defining a few potential objects, but above all varying the modality we wish to study. This approach was applied to the attribute binding in Section~\ref{sec:4_4_attributeBinding} by reducing the number of objects studied and prioritizing the exploration of attribute binding.

Finally, we would like to emphasize that, when exploring semantic links comprehensively, it is essential to test the effect of each word placed together (Section~\ref{sec:catastrophic_neglect}).

\subsection{TIAM with other labels}

TIAM does not depend on the COCO labels and can be applied to other labels as long as we have a detector capable of detecting the desired studied labels. Indeed, TIAM can be implemented with any other detection model, trained on other labels, to evaluate the prompt-image alignment of the T2I models. In particular, the open-vocabulary detection models field has emerged (\eg for detection \cite{gu2022openvocabulary, wu2023cora, li2021grounded} segmentation \cite{li2022languagedriven, ghiasi2022scaling}) presenting itself as a robust contender for surpassing the limitations imposed by constrained label sets.

\subsection{TIAM with other attributes}
See Section~\ref{appendix:other_attributes} of the Supplementary Material for a discussion on how to consider other attributes than \textit{colors}, such as \textit{size} or \textit{texture}.

\end{document}